\documentclass[a4paper,twoside]{article}

\usepackage{epsfig}
\usepackage{authblk}
\usepackage{subcaption}
\usepackage{calc}
\usepackage{amssymb}
\usepackage{amstext}
\usepackage{amsmath}
\usepackage{amsthm}
\usepackage{multicol}
\usepackage{pslatex}
\usepackage{algorithm2e}
\usepackage[bottom]{footmisc}
\usepackage{natbib}
\let\bibhang\relax
\usepackage{apalike}
\usepackage{tabularx}
\usepackage{array}
\usepackage{enumitem}
\usepackage{multirow}
\usepackage{bm}
\usepackage{arydshln}
\usepackage[cal=cm]{mathalfa}
\usepackage{xspace}
\usepackage{xcolor}
\usepackage[colorlinks,linkcolor=red,citecolor=blue,urlcolor=blue]{hyperref}
\usepackage[sort&compress,nameinlink,capitalize]{cleveref}
\usepackage{SCITEPRESS}     

\newcolumntype{C}{>{\centering\arraybackslash}X}
\makeatletter
\DeclareRobustCommand\onedot{\futurelet\@let@token\@onedot}
\def\@onedot{\ifx\@let@token.\else.\null\fi\xspace}

\def\eg{\emph{e.g}\onedot} 
\def\ie{\emph{i.e}\onedot}

\makeatother

\begin{document}

\title{Sensor Generalization for Adaptive Sensing\\in Event-based Object Detection via Joint Distribution Training}
\author{Aheli Saha}
\author{René Schuster}
\author{Didier Stricker}
\affil{Deutsches Forschungszentrum für Künstliche Intelligenz, Kaiserslautern, Germany\\ \texttt{firstName.lastName@dfki.de}}
\keywords{Object Detection, Event Cameras, Adaptive Sensors, Domain Generalization}

\abstract{
Bio-inspired event cameras have recently attracted significant research due to their asynchronous and low-latency capabilities. These features provide a high dynamic range and significantly reduce motion blur. However, because of the novelty in the nature of their output signals, there is a gap in the variability of available data and a lack of extensive analysis of the parameters characterizing their signals. This paper addresses these issues by providing readers with an in-depth understanding of how intrinsic parameters affect the performance of a model trained on event data, specifically for object detection. We also use our findings to expand the capabilities of the downstream model towards sensor-agnostic robustness.}

\onecolumn \maketitle \normalsize \setcounter{footnote}{0} \vfill

\section{\uppercase{Introduction}}
\label{sec:introduction}

Object detection is essential to scene comprehension and forms the basis of increasingly complex visual perception tasks.
Though traditional frame-based object detection has been widely investigated over the years, several challenges remain in highly dynamic environments.
Frame-based cameras capture the entire frame at fixed, regular time intervals. This discrete-time sampling leads to imprecise detection, as fast-moving or transient objects may be completely missed or appear blurry across frames, making it difficult to detect them.
Frame-based cameras also experience substantial latency. Low perceptual latency and accurate and precise decision-making are, however, critical in applications such as autonomous driving. Additionally, static and unchanging components of the scene are captured in each frame, leading to redundant data and requiring high power and bandwidth consumption.

\begin{figure*}
    \centering
    {\setlength{\fboxsep}{7pt}\fbox{\includegraphics[width=\linewidth-15pt]{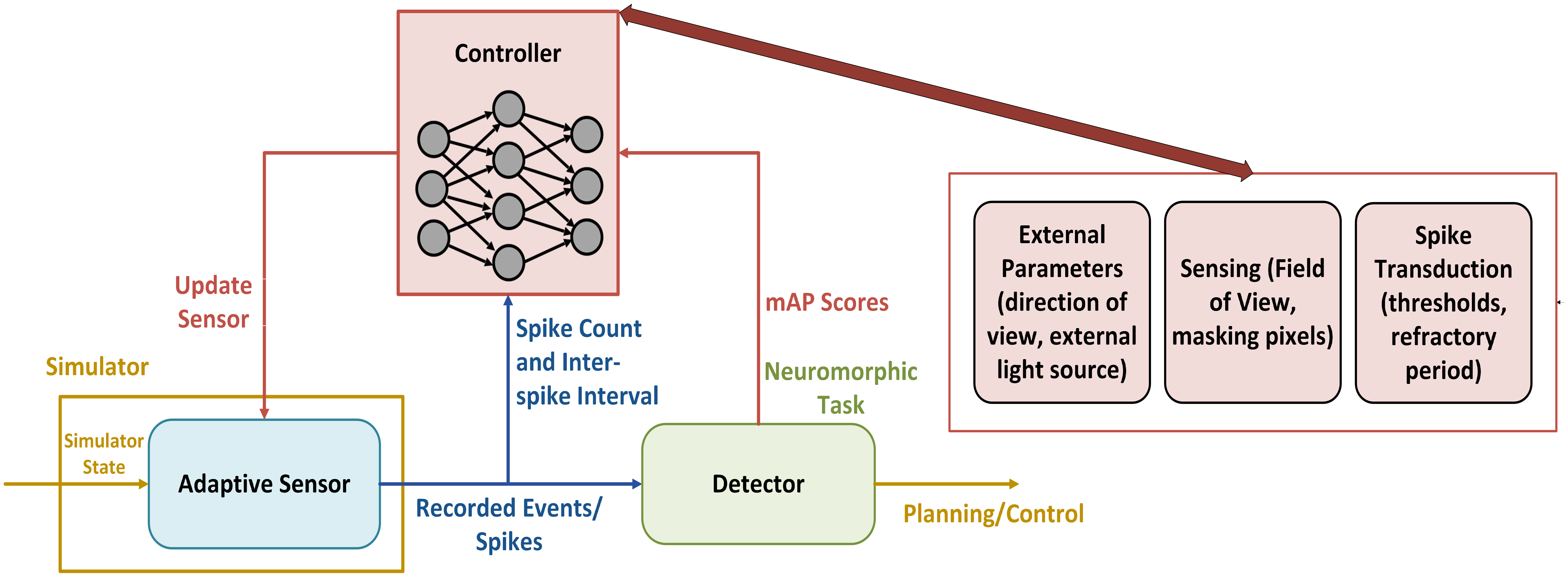}}}
    \caption{Overview of a sensor control loop. We aim to dynamically adapt the sensor characteristics based on energy utilization and task performance feedback. In this work, we investigate and mitigate the undesired effects of dynamically varying spike transduction on task performance, \ie, object detection.}
    \label{fig:control_loop}
\end{figure*}

Event cameras have recently gained considerable significance in this regard. 
They respond to intensity changes at individual pixel positions, triggering an event each time the difference in log intensity crosses a threshold.
The recorded events encode the pixel coordinates, timestamp of occurrence (in millisecond precision), and the binary polarity of the brightness change.
This operational technique allows event cameras to achieve a high dynamic range ($>$ 120 dB) and striking resistance to motion blur. Their asynchronous operation and temporal precision in milliseconds provide unmatched benefits for real-time perception in constantly evolving scenarios.
Due to these developments, event-based object detection has become an expanding field of research (see \cref{sec:related:eventOD}), particularly in the area of autonomous systems, such as drones, vehicles, or robots.
However, to maximize efficiency in size- or power-sensitive applications, it can be argued that sensing and perception need to be considered jointly \citep{eckmann2020active,sprague2015stereopsis}, as is observed in biological organisms to counteract resource constraints. This brings us to the theory of active efficient coding observed in biological organisms, in which the dynamic adaptation of input signals, in conjunction with the environment and task requirements, plays a crucial role. To simulate this evolutionary energy-efficient behavior in artificial systems, we aim to design an adaptive sensor.
The intrinsic and extrinsic characteristics of that sensor can be manipulated by feedback loops to provide optimal information for a certain task in a specific environment, in contrast to a fixed sensor transduction.
Ultimately, this feedback loop is based on performance and energy statistics, which allows for dynamic adjustment for varying environmental stimuli or the current task requirements. A visual representation of this control loop is shown in \cref{fig:control_loop}. Further detailed discussion of this feedback-based dynamic sensor control is beyond the scope of this paper.

However, a key hindrance in the training and functioning of this control loop is the dependence of the detector performance on static input characteristics. A sensor-agnostic detector is therefore a key requirement. In this work, we explore and investigate the challenge of changing event characteristics within an actively controlled sensory-perception system. Modifying the sensor characteristics alters the data distribution of the input signal, which a static, trained event-based object detector is unlikely to be capable of handling. 
To overcome this challenge, we employ a joint-training approach in which our object detector is trained on a set of data from a limited number of different sensors, effectively expanding the known input domain of our model.
The contributions of our paper can be summarized as follows:
\begin{itemize}
    \item We collect an expansive dataset of event data in simulation, with varying sensor characteristics (\cref{sec:method:setup}). To the best of our knowledge, this is the first dataset to probe this dimensionality of event data.
    \item We employ a well-known domain generalization training strategy to expand the model's capabilities to interpolate across any possible sensor configuration in the parameter space.
    \item We design a rigorous exploratory framework and study and report the behavior of the model under distinct experimental test conditions, providing greater insights into the advances and limitations of this method.
\end{itemize}

\section{\uppercase{Related Work}} \label{sec:related}

\subsection{Event-based Object Detection} \label{sec:related:eventOD}

Event-based object detection can be broadly classified into two categories \citep{silva2025recurrent} - (i) direct processing of the sparse data, as in graph neural networks (GNNs) \citep{bi2019graph}, and spiking neural networks (SNNs) \citep{cordone2022object}, and (ii) generation of dense event representations before application of traditional feature extractors.

GNNs use graphs with edges that encode the spatiotemporal connections between nearby events, represented by the nodes \citep{schaefer2022aegnn}.
This preserves the asynchronous, sparse character of event data while enabling localized updates; updates to the pertinent network subgraphs handle new events. While this significantly reduces computations, it is limited by the size and dimensionality of the local subgraphs \citep{gehrig2022pushing}.

The biologically inspired computation of SNNs is particularly well-suited for event data, due to their inherent compatibility with the asynchronous, sparse nature of the data.
Each spiking neuron maintains an internal state and fires when a certain threshold is exceeded, at which stage the signal is propagated to connected neurons \citep{su2023deep,yao2021temporal,zhang2022spiking}.
SNNs maintain low latency and energy efficiency by avoiding static-frame processing and concentrating on dynamic changes.
Compared to regular CNNs, SNNs use significantly less power since they only analyze active events- Spiking-YOLO requires approximately 280x lower energy than Tiny YOLO, \citep{kim2020spiking,su2023deep}.
However, SNNs require surrogate gradients \citep{bendig2023future,su2023deep,guo2023direct,cordone2022object,fan2024sfod} or ANN-to-SNN conversion \citep{bu2022optimized,deng2021optimal,wang2023spike,li2022efficient}, because of the non-differentiable nature of spike generation.
Both of these methods lead to an efficiency-performance trade-off.
Extensive research in this domain is still being conducted, with novel proposals addressing some existing challenges.
EMS-YOLO \citep{su2023deep} enhanced the capacity to directly train deep-SNNs for object identification by creating a full-spike residual block.
SFOD \citep{fan2024sfod} was the first to optimize SNNs for multi-scale feature integration.

Other branches aim to utilize dense feed-forward networks by composing a dense event representation from the sparse data. Some of the most common strategies are Event Histograms \citep{rebecq2017real, maqueda2018event}, Time surfaces \citep{sironi2018hats}, Mixed Density Event Stacks \citep{nam2022stereo}, Time Ordered Event Representations (TORE) \citep{baldwin2022time}, and Event Temporal Images \citep{fan2024dense}. 
Here, we employ the Stacked Histogram Representations as in \citep{gehrig2023recurrent,zubic2024state} in which events in a temporal window are subdivided into smaller temporal bins, spatially distributed according to their pixel positions, and segregated into separate channels by polarity.

Significant progress was achieved with the introduction of benchmark large-scale datasets 1Mpx \citep{perot2020learning} and Gen1 \citep{de2020large}.  They allowed for recent networks like RED \citep{perot2020learning} and ASTMNet \citep{li2022asynchronous}, which use memory methods to leverage the spatiotemporal information in event data fully. RVT \citep{gehrig2023recurrent}, HMNet \citep{hamaguchi2023hierarchical}, and GET \citep{peng2023get} have recently demonstrated outstanding performance employing Transformer networks on dense event representations. 

\subsection{Domain Generalization} \label{sec:related:eventDG}

A domain can be defined as the joint distribution between the input space and the labeled output space. 
Domain generalization (DG) refers to generalizing a model's capabilities to out-of-distribution data that handles unseen test domains \citep{wang2022generalizing,zhou2022domain}. Of the two common and possible settings, single-source DG and multi-source DG, we are particularly interested in the latter, where we have access to multiple similar but distinct source domains. We aim to use this multi-source data to learn feature representations invariant to sensor settings of the source domains.

Current domain generalization research for event data focuses on the adaptability of models trained on limited synthetic data to real-world data, which, owing to non-alignment, results in performance degradation.
The most prevalent source of disparity is the threshold determining the minimum per-pixel intensity fluctuation necessary to trigger an event, which is susceptible to real-time shifts in a real camera \citep{planamente2021da4event}.
Simulators fail to compensate for these non-idealities observed with actual cameras.
To address this, some works like the ones by \citet{stoffregen2020reducing} and \citet{gehrig2020video} operate at the input level, by simulating parameter variations.
DA4Event \citep{planamente2021da4event} proposes to bridge the sim-real gap at the feature level through a multi-view approach employing late fusion over multiple input event representations.
Additionally, models trained for a particular frequency perform poorly at diverging inference frequencies \citep{zubic2024state}. To tackle this issue, FlexEvent \citep{lu2024flexevent} and FAOD \citep{zhang2024frequency} propose modality-fusion and alignment techniques that can leverage both the semantic richness of RGB data and the fine-grained temporal resolution of event data. SSMS \citep{zubic2024state} employs state-space models to train frequency-agnostic models.

Investigating another direction in generalized event cameras, \citet{sundar2024generalized} simulates a single-photon-based sensor design to demonstrate a conceptual design for event cameras capable of adjusting to spatio-temporal contexts.
Although not investigated in this work, it is closely related to our final aim of producing an actively controlled event sensor, and our expanded model would play a significant role in the adaptability of such dynamic models.

Another closely related research direction is domain adaptation for event data, which addresses the scarcity of large-scale labeled event annotations, which impedes their inclusion into mainstream deep learning systems. CTN \citep{zhao2022transformer} leverages the potential of both convolutional modules for early visual processing and Vision Transformers (ViTs) to capture long-range features.
In $\text{DAEC}^2$ \citep{jian2023unsupervised}, various augmentations of an object are projected to a latent space, and an encoder is trained to retain identity across these augmentations. 
However, in this work, we stick to domain generalization across multiple sensor configurations as our main goal, which is a mostly unexplored area in domain generalization, so far.

\section{\uppercase{Sensor-Invariant Training}} \label{sec:method}
\subsection{Data Collection} \label{sec:method:train_data}

\begin{figure*}[t]
  \centering
  \begin{subfigure}[c]{0.32\textwidth}
      \fbox{\includegraphics[width=0.95\linewidth]{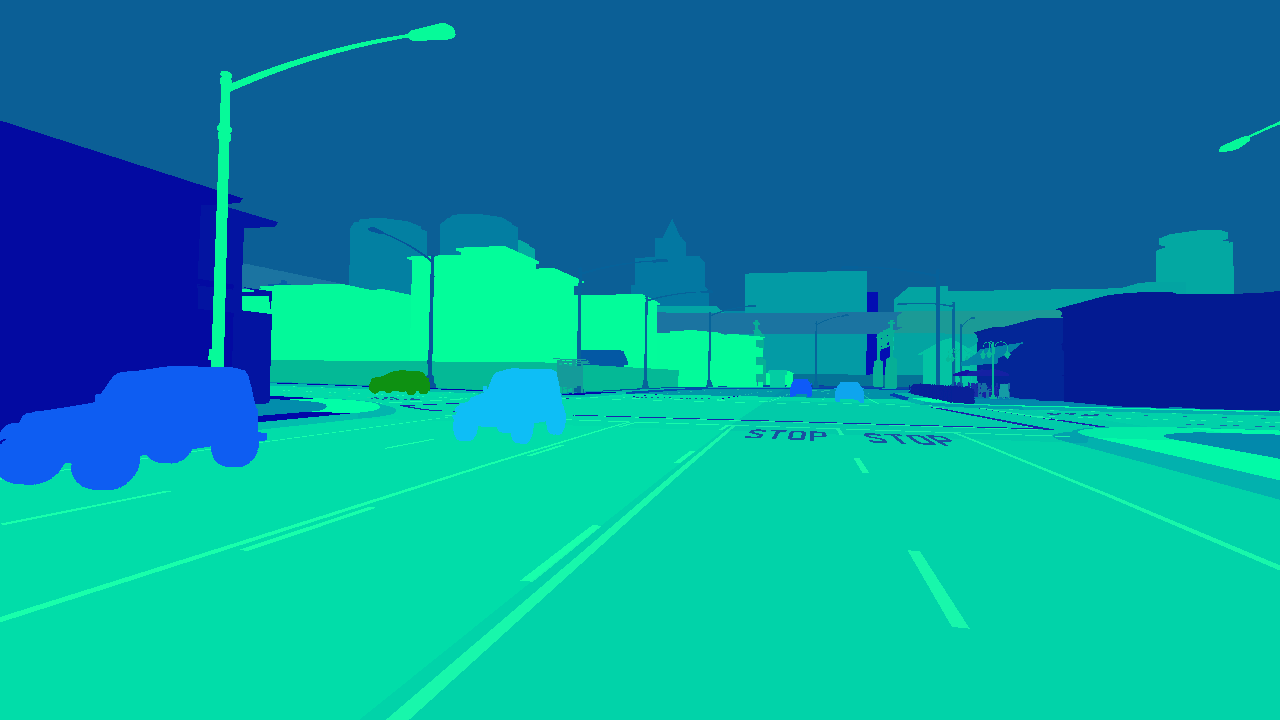}}
  \end{subfigure}
  \vspace{0.5em}
  \begin{subfigure}[c]{0.32\textwidth}
      \fbox{\includegraphics[width=0.95\linewidth]{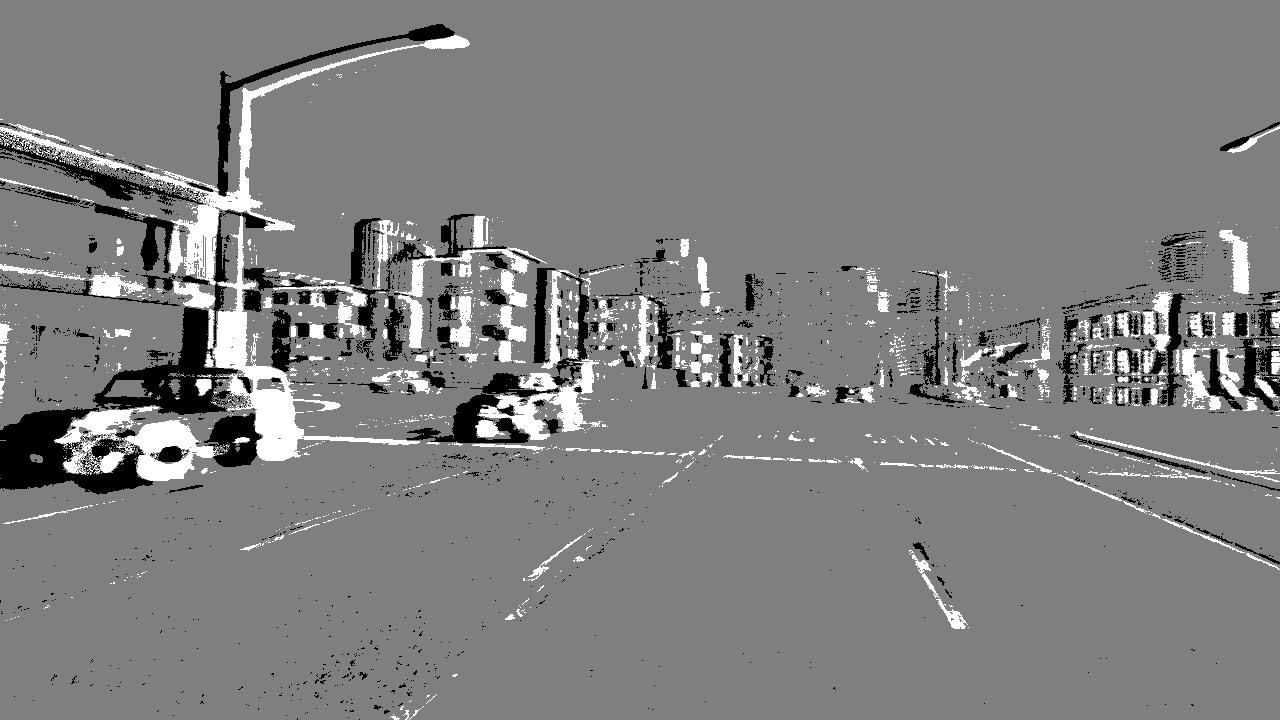}}
  \end{subfigure}
  \begin{subfigure}[c]{0.32\textwidth}
      \fbox{\includegraphics[width=0.95\linewidth]{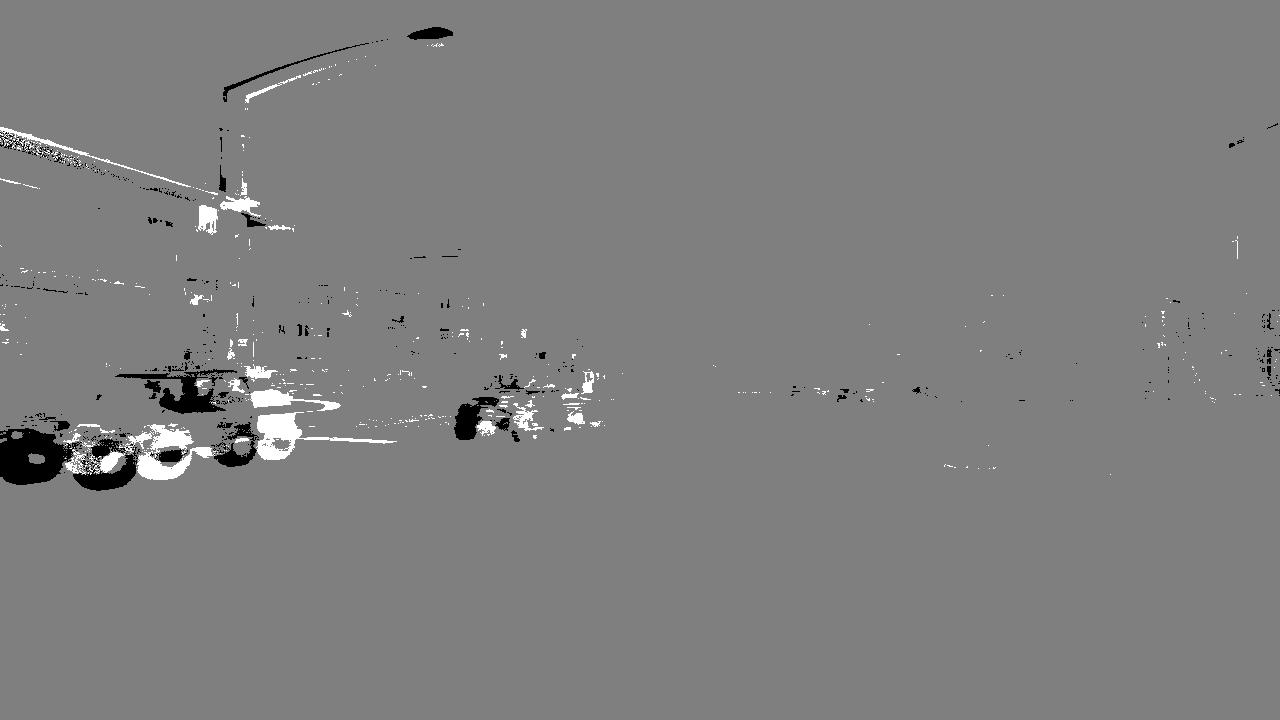}}
  \end{subfigure}
  
  \begin{subfigure}[c]{0.32\textwidth}
      \fbox{\includegraphics[width=0.95\linewidth]{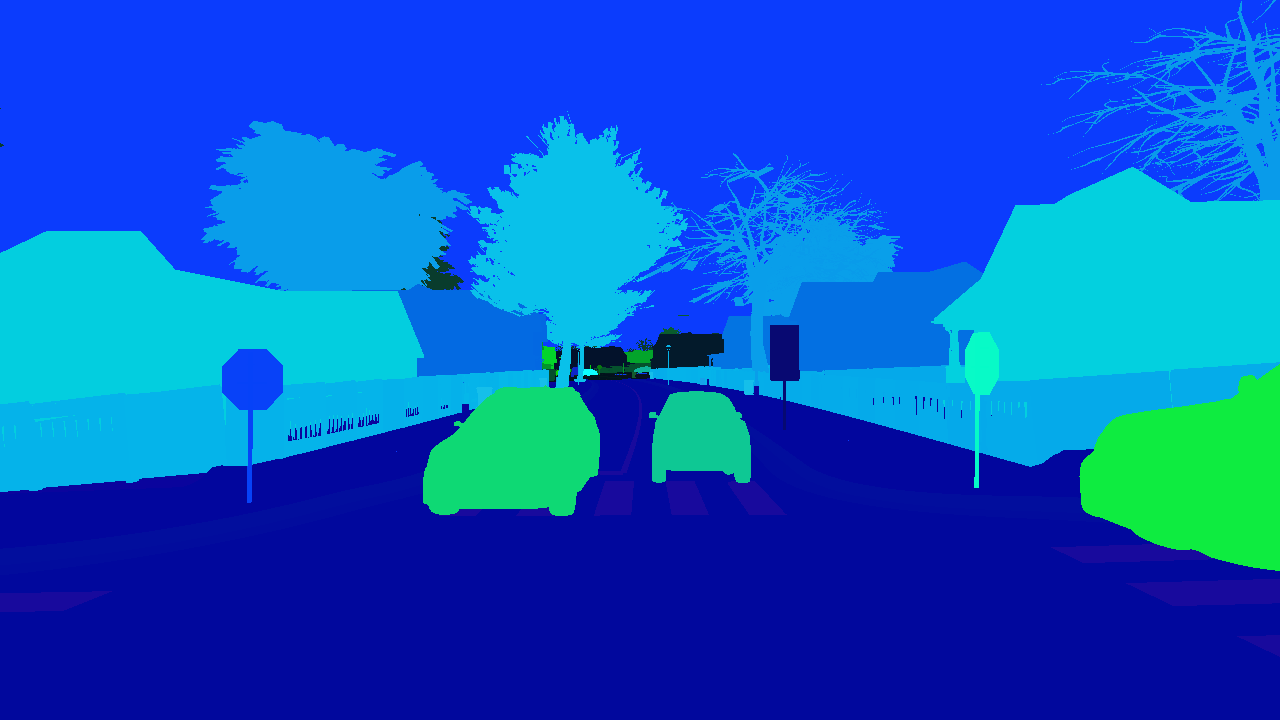}}
      \caption{Instance Segmentation Map}
  \end{subfigure}
  \vspace{0.5em}
  \begin{subfigure}[c]{0.32\textwidth}
      \fbox{\includegraphics[width=0.95\linewidth]{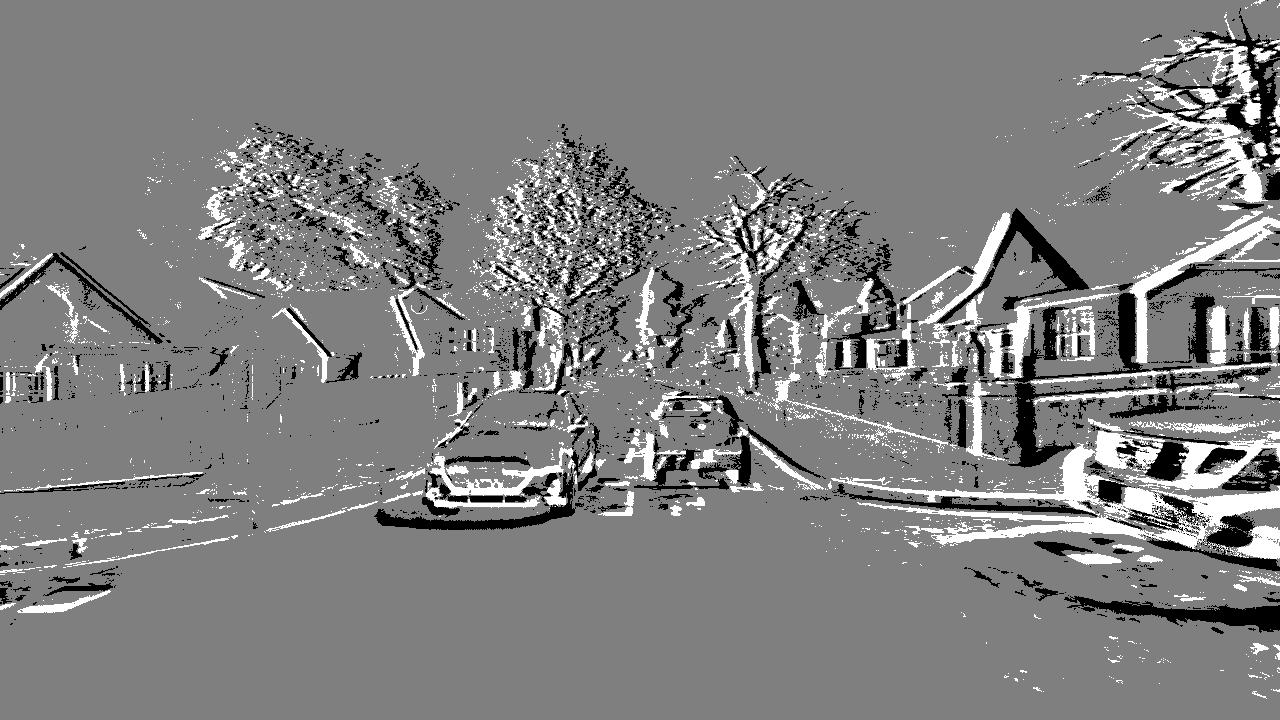}}
      \caption{High Event Density}
  \end{subfigure}
  \begin{subfigure}[c]{0.32\textwidth}
      \fbox{\includegraphics[width=0.95\linewidth]{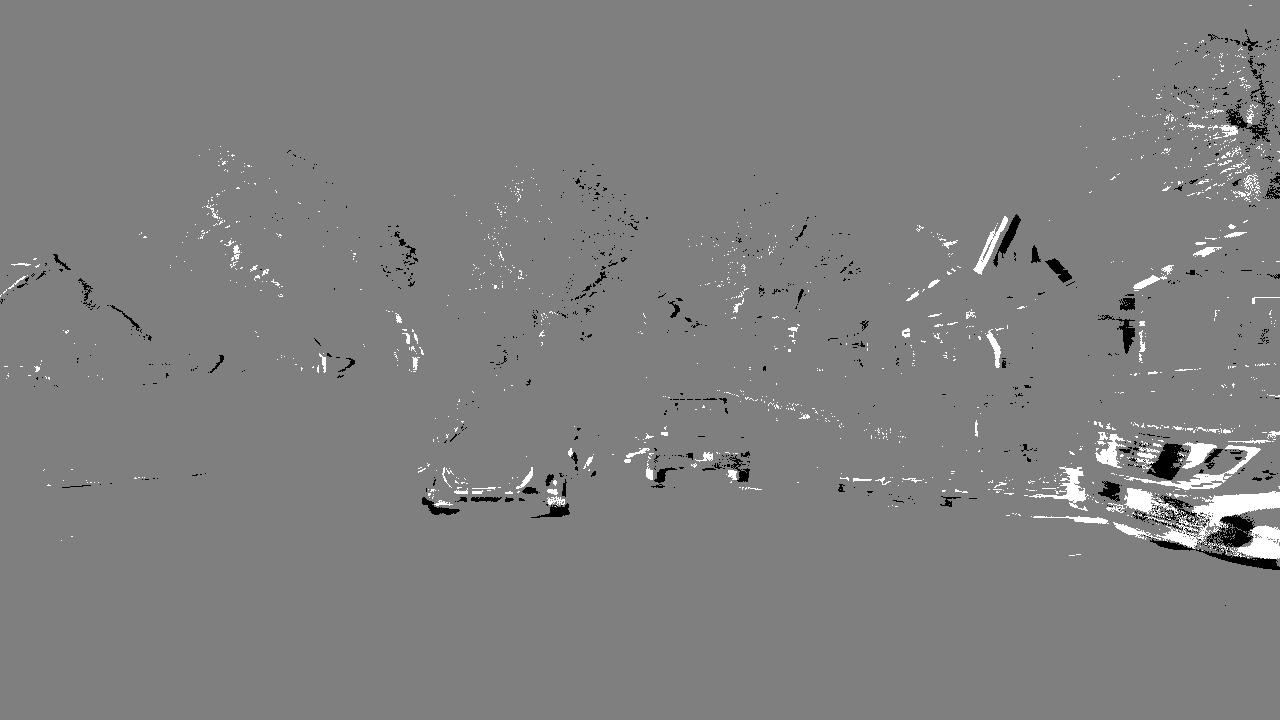}}
      \caption{Low Event Density}
  \end{subfigure}
  \caption{Examples of two scenes with varying configurations which produce high and low event densities, respectively. The black and white points represent negative and positive events, respectively, accumulated over a 50-ms time period. The gray spaces indicate the occurrence of no events within this timeframe at the corresponding pixel locations. The corresponding instance segmentation map is provided for reference to the actual vehicular locations.}
  \label{fig:configs}
\end{figure*}

We use the CARLA simulator \citep{dosovitskiy2017carla} to collect our synthetic datasets. CARLA is pre-equipped with a Dynamic Vision Sensor (DVS) camera, featuring an Application Programming Interface (API) to access its sensory characteristics, which is crucial in our case. Alongside the DVS sensor, we employ the RGB, depth, and instance segmentation sensors on our ego-vehicles to obtain accurate ground-truth bounding boxes for the objects visible in the scene. We retrieve the ground-truth boxes directly from the simulator, while the depth and segmentation modalities are employed to handle occluded or overlapping objects effectively. For demonstration, three classes—cars, buses, and trucks—have been merged into a single vehicle class to increase inter-class variability. Our employed methodology can be seamlessly expanded into multi-class scenarios. 

However, the labels are obtained at a frequency corresponding to that of the RGB camera (\textit{20Hz}), which is significantly lower than that of our event stream, which operates at a sub-millisecond latency. This is a limitation of the simulator. To overcome this, we use temporal windows to accumulate events into corresponding dense representations, such that each window coincides with the next set of labels from the simulator. Each temporal frame is further split into $n_{bins}$ bins and segregated by polarity to produce the corresponding Stacked Histogram Representation, as by \citet{gehrig2023recurrent} and \citet{zubic2024state}, with ($2 \cdot n_{bins} \times H \times W$) tensor dimensions. Each of these tensors corresponds to one labeled output frame.

We collect data by driving our ego-vehicle across 12 distinct pre-determined routes in each of 13 different town maps, to produce variability in traffic density and weather conditions. 500 frames are collected for each sequence, at a frame rate of \textit{20Hz}, providing us with 65 minutes of recorded data. Note that this frame rate refers to that of the traditional sensors, and only our event frames are synchronized to these. 
All events triggered in the corresponding $0.05s$ temporal window are recorded with their accurate and precise timestamps in \textit{nanoseconds}.
This produces high-quality, high-resolution (720 x 1280) data on driving environments, spread across diverse environmental scenarios and sensor characteristics.  We also filter out bounding boxes with a side length lower than 20 pixels or a diagonal length below 60 pixels, following the evaluation criteria of \citet{perot2020learning}. Contrary to the 1Mpx dataset, we do not downsample our data and use the original high resolution. Towns are split for train, validation, and test, using a 70-15-15\% split, resulting in $(9,2,2)$ towns for each, respectively.

This collection procedure is repeated over 14 distinct DVS-sensor settings, outlined in \cref{table:data_config} and discussed further in \cref{sec:method:setup}.
Our dataset, of $\sim15$ hours in total, thus provides a significant benchmark for numerous baseline and comparative studies on sensor behavior, enabling robust and sensor-invariant models for downstream tasks, and the study and development of dynamically adaptive event sensors.

\subsection{Experimental Setup} \label{sec:method:setup}

\subsubsection{\textbf{Parameter Space}}
We focus on four key parameters that significantly influence event camera behavior - the intensity thresholds for event triggering (one for each polarity), the minimum threshold for interspike interval, and the field of view. We define our set of parameters $\mathcal{P}$ as:
\begin{align}
    \mathcal{P} = \{th_p, th_n, T_r, F_v\}
\end{align}
where:
\begin{itemize}[leftmargin=*]
    \item $th_p$ represents the event threshold for positive events, \ie an increase in logarithmic intensity, controlling the sensitivity of event generation.
    \item $th_n$ represents the event threshold for negative occurrences, thus also adapting sensor sensitivity.
    \item $T_r$ denotes the refractory period, affecting the temporal resolution of events.
    \item $F_v$ stands for the field of view (FoV), determining the angular extent of the sensor coverage.
\end{itemize}

\begin{table}[t]
        \caption{Data Configurations.}
        \label{table:data_config}
        \renewcommand{\arraystretch}{1}
        \begin{tabularx}{\linewidth}{C|C|C|C|C}
            \hline\hline
            \textbf{Config.} (${\mathbfcal{E}}_{\bm{i}}$) & \textbf{Pos. Thr.} ($\bm{th_p}$) & \textbf{Neg. Thr.} ($\bm{th_n}$) & \textbf{Ref. Period} ($\bm{T_r}$) [ms] & \textbf{Field of View} ($\bm{F_v}$) \\ 
            \hline
            \hline
            $\mathbfcal{E}_{\bm{base}}$ & 0.5 & 0.5 & 10 & $90^\circ$ \\ 
            \hline
            $\mathbfcal{E}_{\bm{1}}$ &  \textbf{0.25} & \textbf{0.25} & 0.01 & $90^\circ$ \\ 
            $\mathbfcal{E}_{\bm{2}}$ & \textbf{0.75} & \textbf{0.75} & 0.01 & $90^\circ$ \\ 
            $\mathbfcal{E}_{\bm{3}}$ & \textbf{1.0} & \textbf{1.0} & 0.01 & $90^\circ$ \\ 
            $\mathbfcal{E}_{\bm{4}}$ & 0.5 & 0.5 & \textbf{10} & $90^\circ$ 
            \\ 
            $\mathcal{E}_{\bm{5}}$ & 0.5 & 0.5 & \textbf{25} & $90^\circ$ \\ 
            $\mathbfcal{E}_{\bm{6}}$ & 0.5 & 0.5 & \textbf{50} & $90^\circ$ \\ 
            $\mathbfcal{E}_{\bm{7}}$ & 0.5 & 0.5 & 0.01 & ${45^\circ}$ \\ 
            $\mathbfcal{E}_{\bm{8}}$ & 0.5 & 0.5 & 0.01 & ${135^\circ}$ \\ 
            $\mathbfcal{E}_{\bm{9}}$ & 0.5 & 0.5 & 0.01 & ${160^\circ}$ \\ 
            \hline
            $\mathbfcal{E}_{\bm{10}}$ & 0.25 & 0.25 & 50 & $45^\circ$ \\ 
            $\mathbfcal{E}_{\bm{11}}$ & \textbf{1.0} & \textbf{0.5} & 25 & $90^\circ$ \\ 
            $\mathbfcal{E}_{\bm{12}}$ & 0.7 & 0.7 & 20 & $65^\circ$ \\ 
            $\mathbfcal{E}_{\bm{13}}$ & \textbf{0.3} & \textbf{0.9} & 15 & $130^\circ$ \\ 
            \hline
            \hline
        \end{tabularx}
\end{table}

\subsubsection{\textbf{Configurations}} \label{sec:results:setup:configs}
We define a comprehensive set of experimental configurations $\mathcal{E} = \{\mathcal{E}_{base}, \mathcal{E}_1, \ldots, \mathcal{E}_{13} \}$, where each configuration $\mathcal{E}_i \in \mathcal{E}$ represents a unique combination of event camera parameters. For any configuration $\mathcal{E}_i$ and parameter $p \in \mathcal{P}$, we denote the value of $p$ in $\mathcal{E}_i$ as $p(\mathcal{E}_i)$.

We define a standard or default configuration, referred to as $\mathcal{E}_{base}$, which we use as the baseline to adapt the parameters in the subsequent configurations. This is the default sensor configuration provided by CARLA, except for the refractory period, which has been set to a very low value (0.01ms) instead of the default value of 0. In configurations $\mathcal{E}_i, i \in \{ 1,2,..,9 \} $, we focus on varying a single setting $p \in \mathcal{P}$ over 3 distinct values for each, producing a distribution over each parameter space. At this stage, we do not consider $th_p(\mathcal{E}_i), th_n(\mathcal{E}_i)$ independently of each other, and vary them simultaneously. For settings $\mathcal{E}_i, i \in \{ 10, 11, 12, 13 \}$, we fluctuate all the parameters simultaneously to generate unique distributions over the parameter space. Specifically for $\mathcal{E}_{11}$ and $\mathcal{E}_{13}$ we consider each of $th_p(\mathcal{E}_i), th_n(\mathcal{E}_i)$ independently, and set them at contrasting values. The specific parameter values for each configuration are detailed in \cref{table:data_config} and visual examples for some of the configurations are displayed in \cref{fig:configs}.

\subsection{Dataset Partitioning}

\begin{figure}
    \centering
    \includegraphics[width=\linewidth]{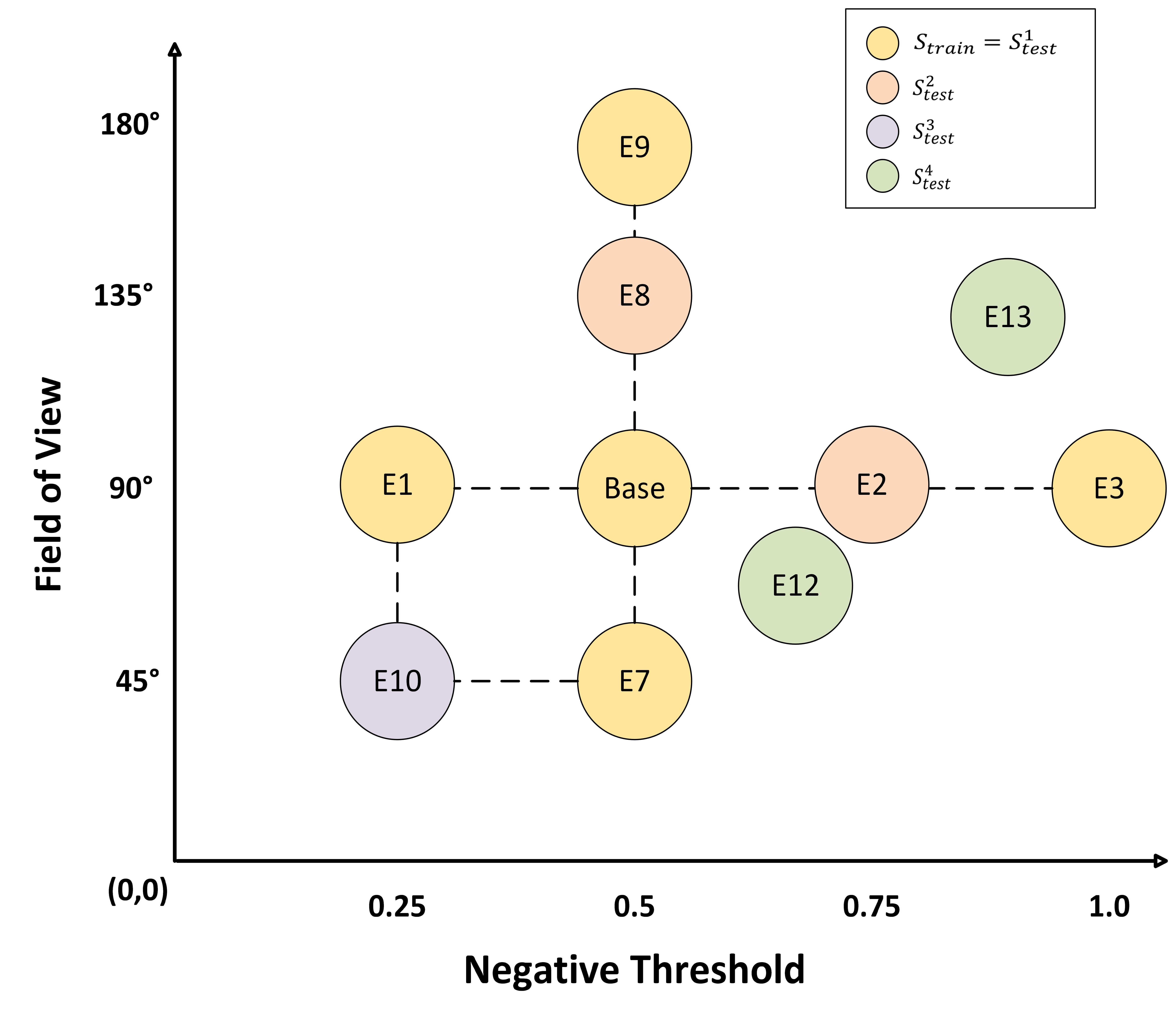}
    \caption{Simplified overview of the design and positioning of the distinct test sets in the parameter space of two parameters - the field of view and negative threshold. The training set and the first test set consist of distinct configurations varying along each dimension. The second test set consists of configurations obtained by interpolating along each dimension, whereas the third set is obtained by combining seen parameters into unique configurations. The fourth set represents the unseen parameter space for each parameter.}
    \label{fig:test_configs_overview}
\end{figure}

\label{sec:method:partitioning}
Our experimental framework is designed to rigorously assess the generalization capabilities of event-based vision models across various event camera configurations. By leveraging the synthetic dataset generated in a controlled simulated environment, we partition our data into distinct training and test sets to design a systematic evaluation of multiple aspects of generalization. \cref{fig:test_configs_overview} provides a simplified visual overview of the main design idea.

\begin{enumerate}[wide,labelindent=0pt,itemsep=8pt,listparindent=0pt]
    \item Training Set:
    
     This set, denoted by $S_{train}$, encompasses a diverse range of parameter combinations, serving as the foundation for model training and establishing the baseline performance. The idea is to use the lower and higher bounds of the range for each parameter $ p_i \in \mathcal{P} $ for training to expose the model to the extreme variations in the data distribution. This is supplemented by $base$ configuration, $\mathcal{E}_{base}$, on top of which the variations are levied. We hypothesize that by learning the patterns of the limits of the data, the model can better adapt to unseen intermediate values. We compose this set as:
    \begin{align}
        S_{train} = \{ \mathcal{E}_{base}, \mathcal{E}_1, \mathcal{E}_3, \mathcal{E}_4, \mathcal{E}_6, \mathcal{E}_7, \mathcal{E}_9 \}   
    \end{align}

    \item Test Sets:
    
    We define four test sets, each designed to evaluate a specific aspect of model generalization:

    \begin{enumerate}[wide,labelindent=0pt,itemsep=8pt,listparindent=0pt]
        \item Intra-distribution Generalization:
        
        We use the same sensor characteristics for this test set as are used by our training set $S_{train}$. By employing the same parameter distribution, but a different split of the data (\eg different towns), this approach assesses the model's ability to generalize within the training parameter space. It is aimed at establishing a baseline performance and demonstrating the model's robustness to external fluctuations, like geographical features, lighting conditions, and traffic density, within familiar non-divergent settings. Defined as:
        \begin{align}
            S_{test}^1 
            &= S_{train}
        \end{align}

        This train-test split is representative of what is commonly used in the literature, \ie, you have data from a single source and partition it.
        Please note that $\mathcal{E}_i$ describes the sensor configuration only; the training and test splits are generated for distinct towns. 
        
        \item Single Parameter Perturbation:
        
        In this setup, each configuration is deliberately designed to differ from all training configurations by exactly one parameter. This focused approach is aimed at isolating the influence each parameter has on the model's performance by eliminating confounding factors. It is crucial for uncovering targeted insights into the model's sensitivity, which can guide further model refinement.
        Furthermore, we can investigate if the model indeed learns to interpolate a single parameter between the provided limits during training. Formally:
        \begin{align}
            S_{test}^2 
            &= \Big\{ \mathcal{E}_2, \mathcal{E}_5, \mathcal{E}_8 \Big\}
        \end{align}
        
        \item Distinct Configuration Derived from Observed Parameters:
        
        Each parameter is sampled from the well-defined training distribution, but their combined state maps them to a novel resulting distribution. It aims to assess the model's potential to interpolate between familiar configurations, demonstrating the model's understanding of relationships between parameters, instead of overfitting to a singular parameter space. Evaluating its capacity to handle the complex sensor dynamics resulting from the interplay of multiple internal factors is essential for the final use case, as each parameter can be adjusted independently by a controller, which would produce unique distributions in the combined space.
        Formally put:
        \begin{align}
            S_{test}^3 
            &= \Big\{ \mathcal{E}_{10}, \mathcal{E}_{11} \Big\}
        \end{align}

        \item Arbitrary Combinations of Unseen Parameters:
        
        This test evaluates the model's ability to perform beyond the known parameter values of the training data by employing entirely unseen individual parameters. It aims to assess the model's capability and robustness when operating under significant shifts in the event camera's operating characteristics. This is especially relevant for our ultimate objective of developing a dynamically adaptive event sensor, since it would theoretically be possible for the sensor to assume any permitted state. Ensuring the downstream model performs optimally, invariant to the sensor characteristics, is critical.
        Formally defined as:
        \begin{align}
            S_{test}^4 
            &= \Big\{ \mathcal{E}_{12}, \mathcal{E}_{13} \Big\}
        \end{align}
    \end{enumerate}
\end{enumerate}

\subsection{Network Architecture} \label{sec:method:model}

We use two off-the-shelf event-based object detectors, Recurrent Vision Transformers (RVT) \citep{gehrig2023recurrent} and State Space Models (SSMs) \citep{zubic2024state}, for our experiments and evaluation. An overview of their concepts and architectures is discussed here.

\subsubsection{Recurrent Vision Transformers (RVT)} 
The proposed architecture \citep{gehrig2023recurrent} employs a combination of convolutional layers, self-attention, and LSTMs to preserve and extract spatio-temporal information from dense tensor representations of the events. 

Overlapping convolutional kernels are used to embed spatial relations between feature positions and for downsampling output maps of preceding layers. Self-attention is applied in two stages to cater to both local feature interactions using block attention and global feature interactions using grid attention. This strategy of global dilation of features counteracts the quadratic complexity that would have been otherwise required to apply self-attention over the entire feature map. Features across time-steps are aggregated with LSTMs initialized with states from the preceding time-steps. 

This architecture block is repeated four times sequentially, and the hidden states of the LSTM, from the second to the fourth layer, are employed as features for the object detection head.

\subsubsection{State Space Models (SSMs)} 
The key characteristic of \citep{zubic2024state} involves replacing the LSTM blocks in the above-discussed architecture with State Space Model layers. This allows for faster parallel training utilizing convolutions and achieves a $33\%$ more rapid training than employing LSTMs, by avoiding non-linearities which necessitated sequential processing of the data. 

It also addresses the issue of performance degradation when employed at frequencies (temporal windows) divergent from the training frequency. Learnable timescale parameters are introduced, along with anti-aliasing strategies when the kernel bandwidth exceeds the Nyquist frequency. 

Additionally, at higher sampling frequencies (beyond the Nyquist frequency), aliasing can severely impact performance. This is handled by a frequency-selective masking where the kernel focuses on lower frequencies, smoothening high-frequency components surpassing an empirical threshold. An additional $H_2$ norm regularization suppresses the frequency response of the system beyond a range given by $[w_{min}, w_{max}]$, during the computation of the norm of the system's transfer function.

\subsection{Implementation Details} \label{sec:method:implementation}

We train and compare the performance of our joint training set against a base model for both the above architectures. All models are trained with a 32-bit precision over 400k steps, and employ the Adam optimizer and the OneCycle learning rate scheduler as in \citet{gehrig2023recurrent,zubic2024state}. The best-performing checkpoint on the validation dataset is used for testing across all scenarios.

Random horizontal flips and zoom-in, zoom-out procedures are used to supplement the data. For the event-based representations, we utilize fixed temporal windows of \textit{50ms}, and discretize them over $n_{bins}=10$ bins. We use a batch size of 8, a sequence length of 5, and a global learning rate of $2e^{-4}$ for all our experiments. The YOLOX \citep{ge2021yolox} detection head incorporates an IOU loss, along with the regression loss, which is aggregated over batches and sequences in each step. The $H_2$ norm is additionally incorporated with the loss for the models using SSMs.

\section{\uppercase{Experiments and Results}} \label{sec:results}

We present and discuss our results in this subsection, providing a comprehensive analysis of the model's performance across the previously designed test settings. We employ standard evaluation metrics to quantify and compare all models, providing a leveled ground for further analysis.

\subsection{Evaluation Framework}
\label{sec:results:metrics}

For each test set $S_{test}^k$, we compute a set of performance metrics $\mathcal{M} = \{ m_1, m_2, ..., m_n \} $. We use the COCO evaluation metrics \citep{lin2014microsoft} for object detection, where the Average Precision (AP) is the primarily employed metric. We define $\mathcal{M} = \{ \text{AP, } \text{AP}_{50}, \text{ AP}_{75}, \text{ AP}_{L}, \text{ AP}_{M} \}$. Here, AP is used as a shorthand for AP@[0.50:.95], which reports the average over IoU thresholds from 0.50 to 0.95, and $\text{AP}_{50}$ and $\text{AP}_{75}$ are the AP scores for IoU values of 0.50 and 0.75, respectively. $\text{AP}_M$ and $\text{AP}_L$ refer to the scores for medium ($32 < \text{side length} < 96$ pixels) and large objects ($\text{side length} > 96$ pixels), respectively. We aim to provide an extensive knowledge of the model's generalization capabilities, strengths, and limits in the context of event-based vision tasks across many event camera configurations by methodically investigating these performance indicators throughout our carefully prepared test sets.

 The performance of a model $f$ on $S_{test}^k$ is defined:
\begin{align}
    \text{Score}_k(f) = \{ \mathbb{E}_{\mathcal{E}_i \in S_{test}^k}[m_j(f, \mathcal{E}_i)] \mid m_j \in \mathcal{M} \}
\end{align}

where $m_j(f, \mathcal{E}_i)$ denotes the performance of model $f$ on configuration $\mathcal{E}_i$ according to metric $m_j$.

\subsection{Sensor Generalization} \label{sec:results:generalization}

\begin{figure*}[t]
  \centering
  \hspace*{0.02\textwidth}
  \begin{subfigure}[c]{0.32\textwidth}
      \centering \small Sample from $\mathcal{E}_{base}$
  \end{subfigure}%
  \begin{subfigure}[c]{0.32\textwidth}
      \centering \small Sample from $\mathcal{E}_{3}$
  \end{subfigure}%
  \begin{subfigure}[c]{0.32\textwidth}
      \centering \small Sample from $\mathcal{E}_{8}$
  \end{subfigure}
  \vspace{0.25em}
  
  \begin{subfigure}[c]{0.02\textwidth}
    \rotatebox[origin=c]{90}{\small RVT-B ($\mathcal{E}_{base}$)}
  \end{subfigure}
  \begin{subfigure}[c]{0.32\textwidth}
      \centering
      \fbox{\includegraphics[width=0.95\linewidth]{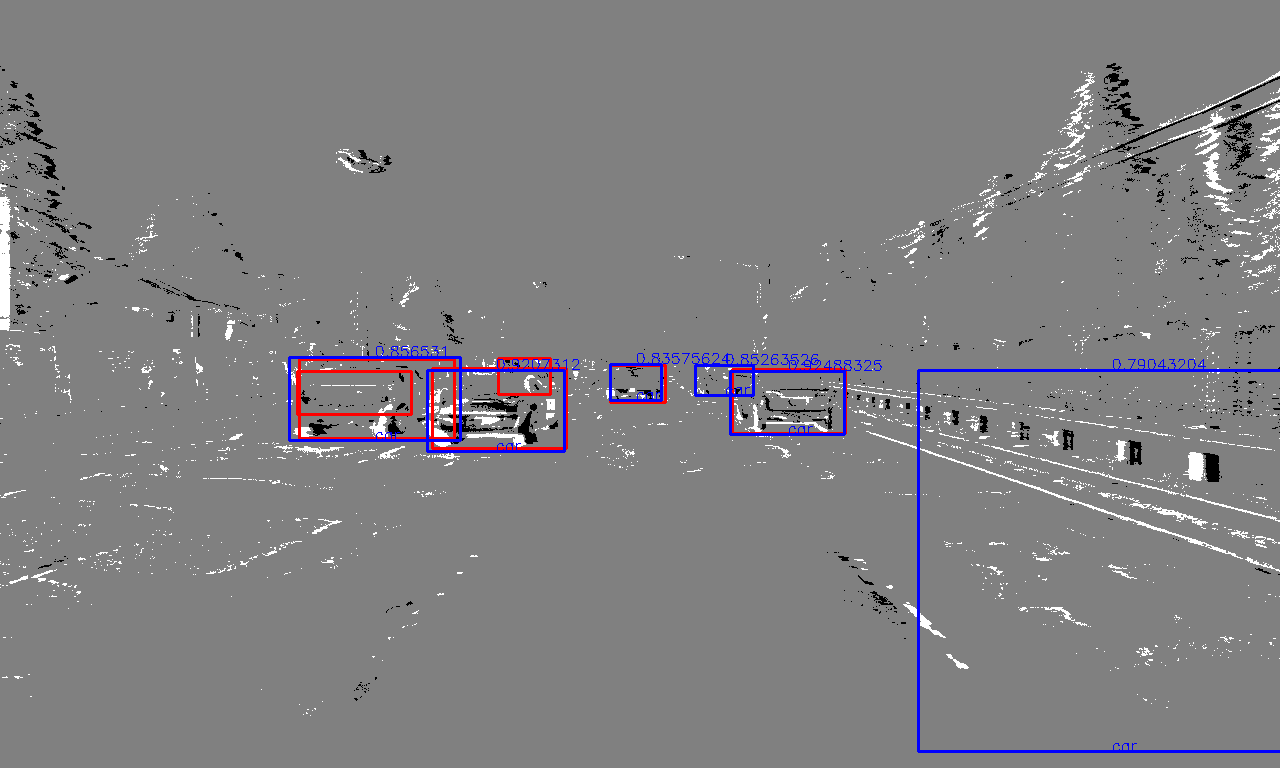}}
  \end{subfigure}
  \begin{subfigure}[c]{0.32\textwidth}
      \centering
      \fbox{\includegraphics[width=0.95\linewidth]{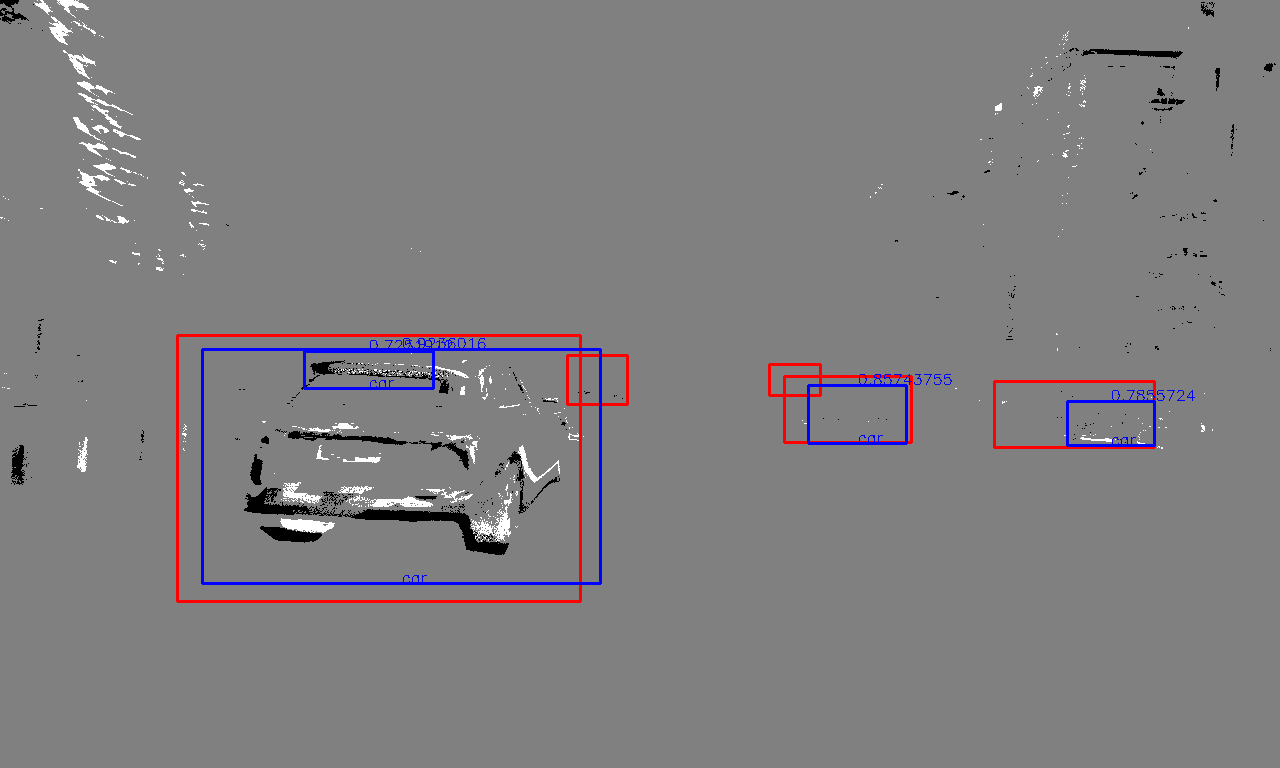}}
  \end{subfigure}
  \begin{subfigure}[c]{0.32\textwidth}
      \centering
      \fbox{\includegraphics[width=0.95\linewidth]{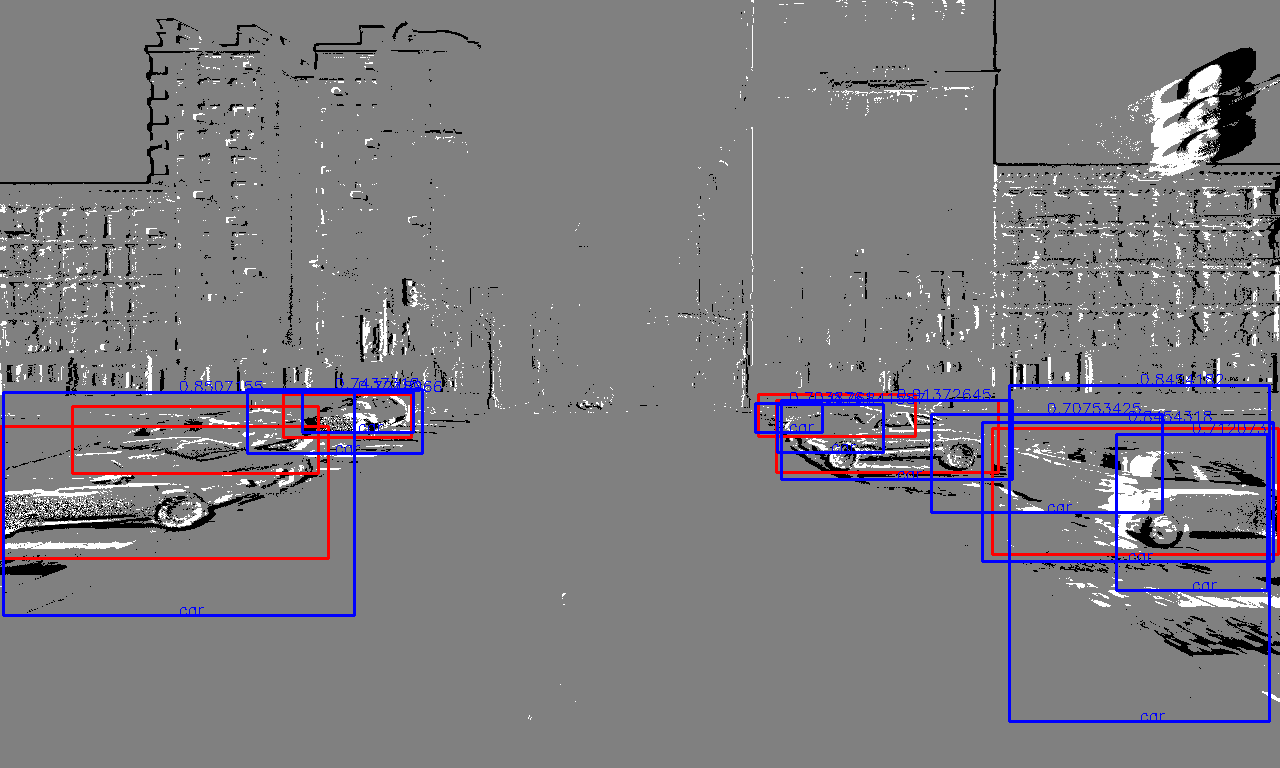}}
  \end{subfigure}
  \vspace{0.5em}
  
  \begin{subfigure}[c]{0.02\textwidth}
    \rotatebox[origin=c]{90}{\small RVT-B ($S_{train}$)}
  \end{subfigure}
  \begin{subfigure}[c]{0.32\textwidth}
      \centering
      \fbox{\includegraphics[width=0.95\linewidth]{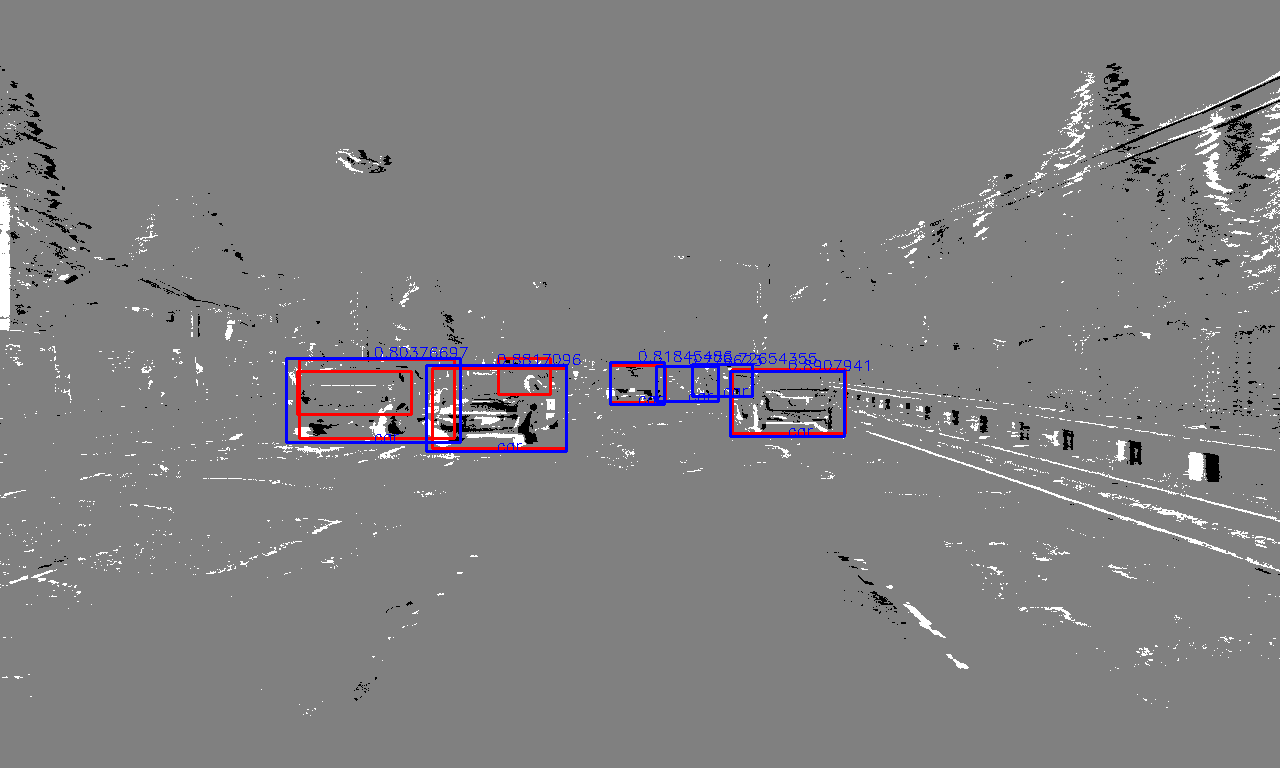}}
  \end{subfigure}
  \begin{subfigure}[c]{0.32\textwidth}
      \centering
      \fbox{\includegraphics[width=0.95\linewidth]{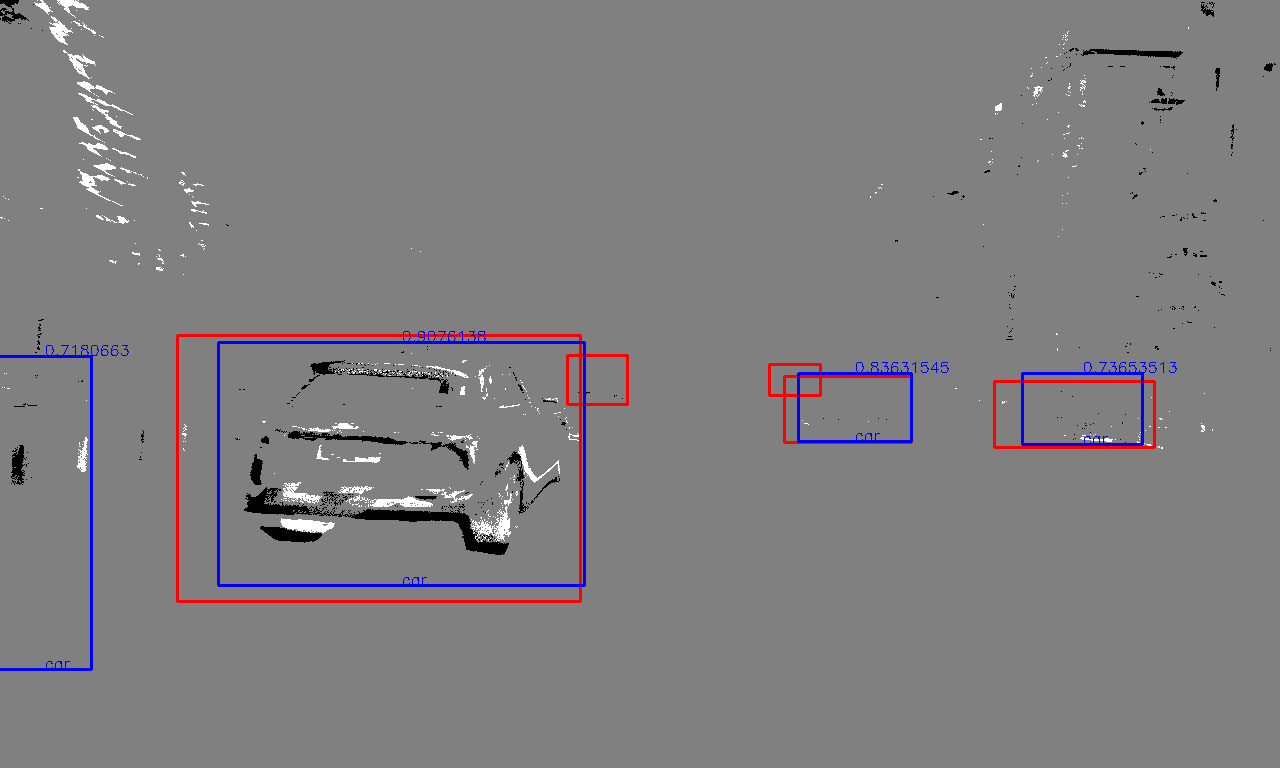}}
  \end{subfigure}
  \begin{subfigure}[c]{0.32\textwidth}
      \centering
      \fbox{\includegraphics[width=0.95\linewidth]{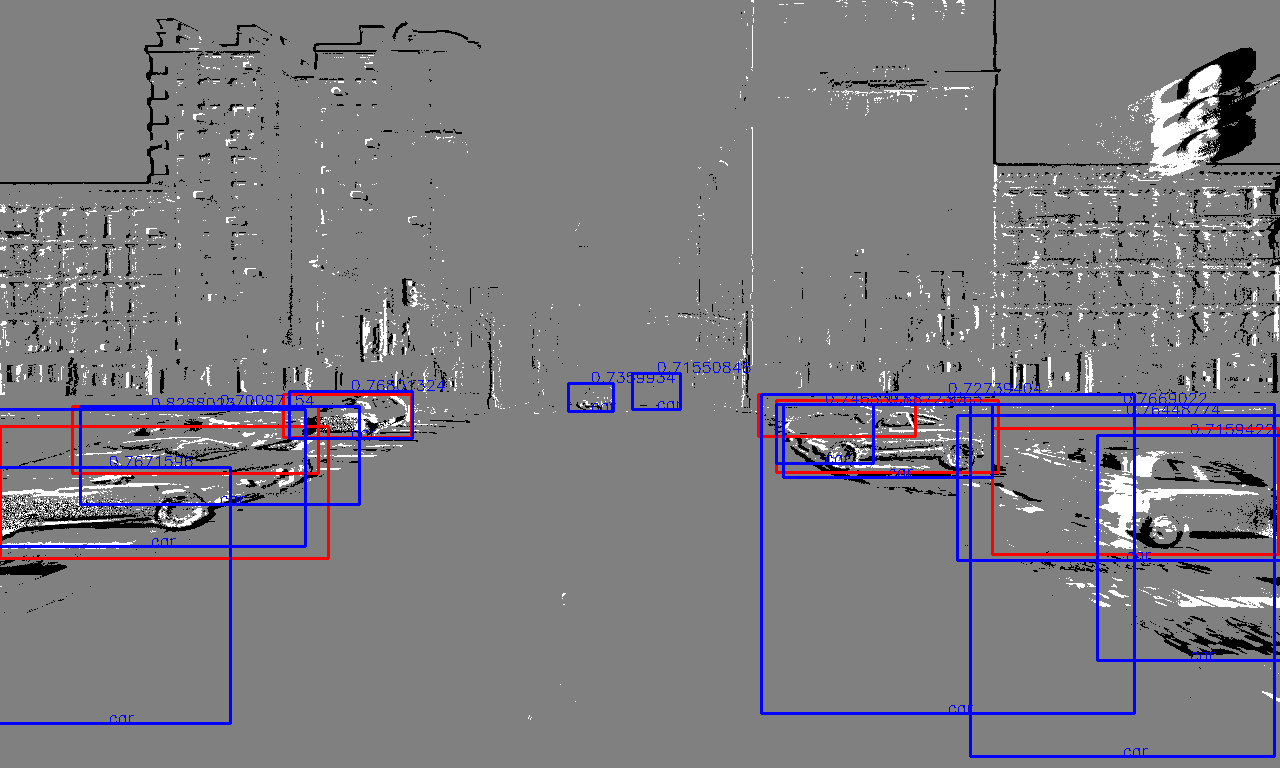}}
  \end{subfigure}
  \vspace{0.5em}
  
  \begin{subfigure}[c]{0.02\textwidth}
    \rotatebox[origin=c]{90}{\small SSMS-B ($\mathcal{E}_{base}$)}
  \end{subfigure}
  \begin{subfigure}[c]{0.32\textwidth}
      \centering
      \fbox{\includegraphics[width=0.95\linewidth]{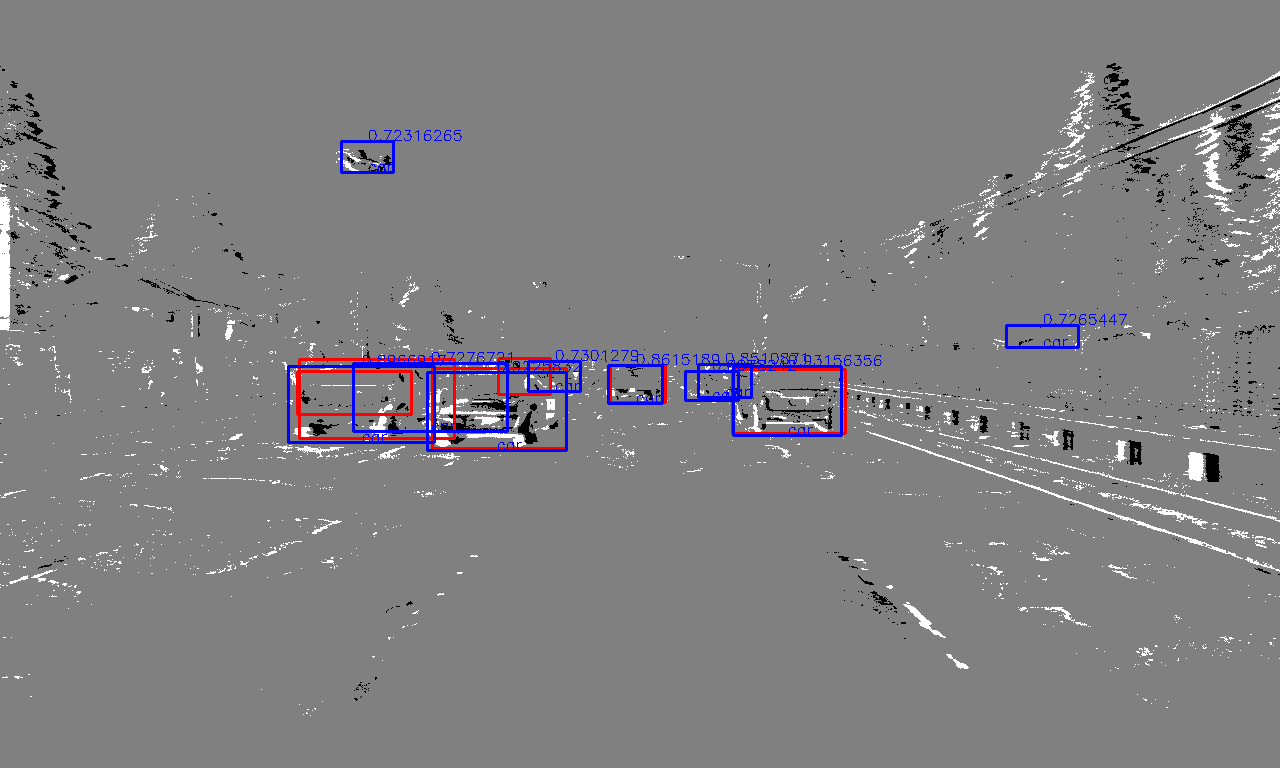}}
  \end{subfigure}
  \begin{subfigure}[c]{0.32\textwidth}
      \centering
      \fbox{\includegraphics[width=0.95\linewidth]{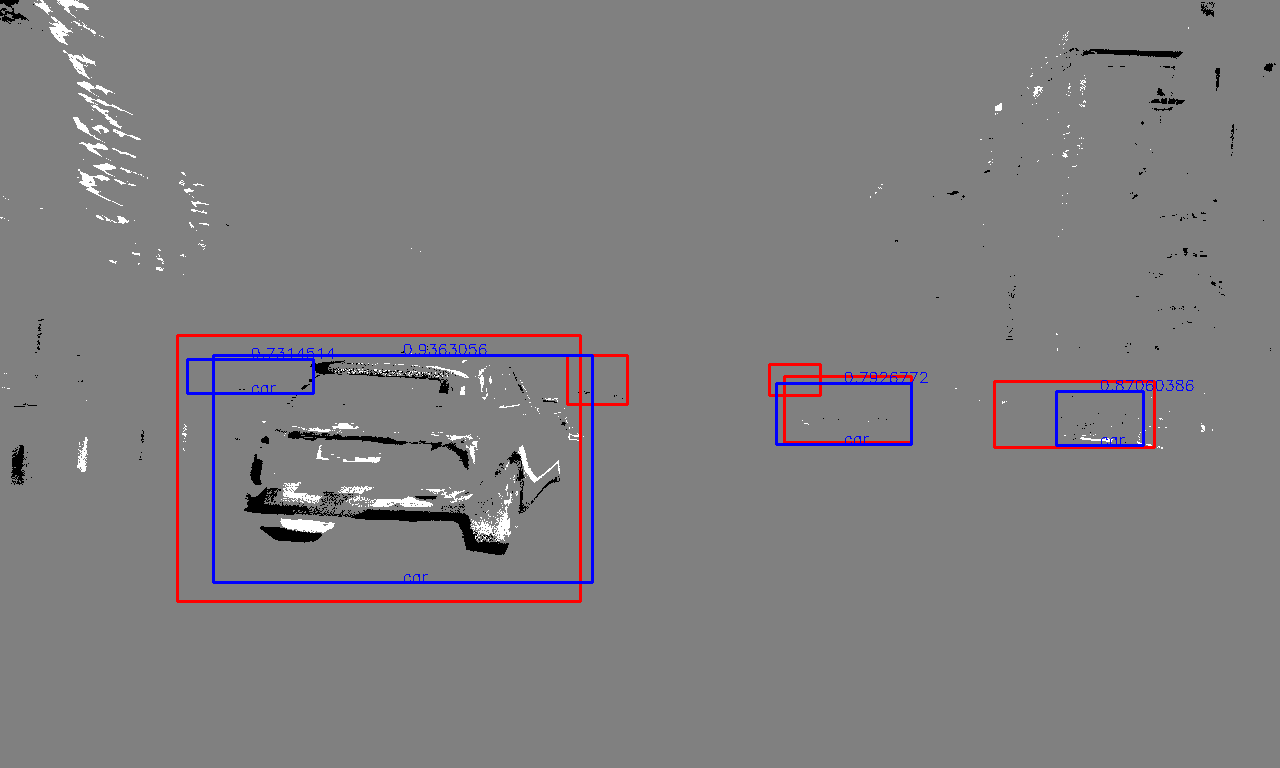}}
  \end{subfigure}
  \begin{subfigure}[c]{0.32\textwidth}
      \centering
      \fbox{\includegraphics[width=0.95\linewidth]{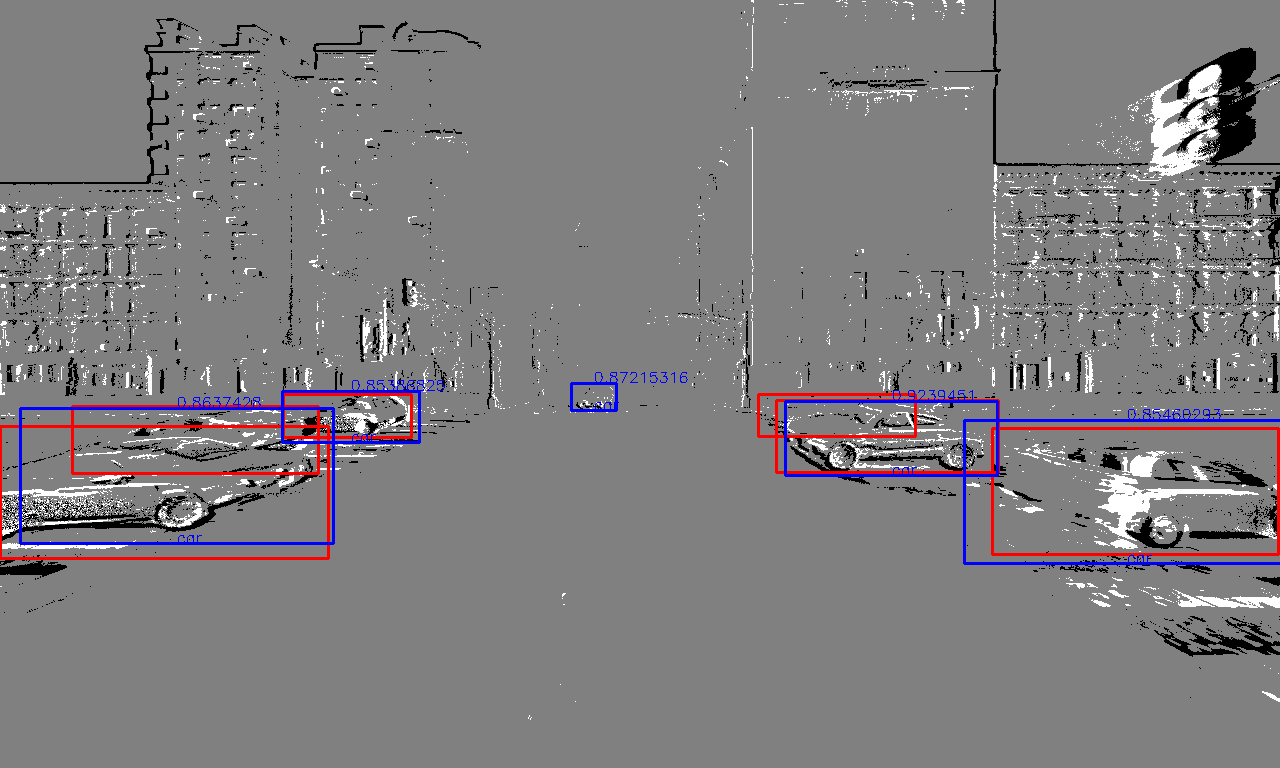}}
  \end{subfigure}
  \vspace{0.5em}
  
  \begin{subfigure}[c]{0.02\textwidth}
    \rotatebox[origin=c]{90}{\small SSMS-B ($S_{train}$)}
  \end{subfigure}
  \begin{subfigure}[c]{0.32\textwidth}
      \centering
      \fbox{\includegraphics[width=0.95\linewidth]{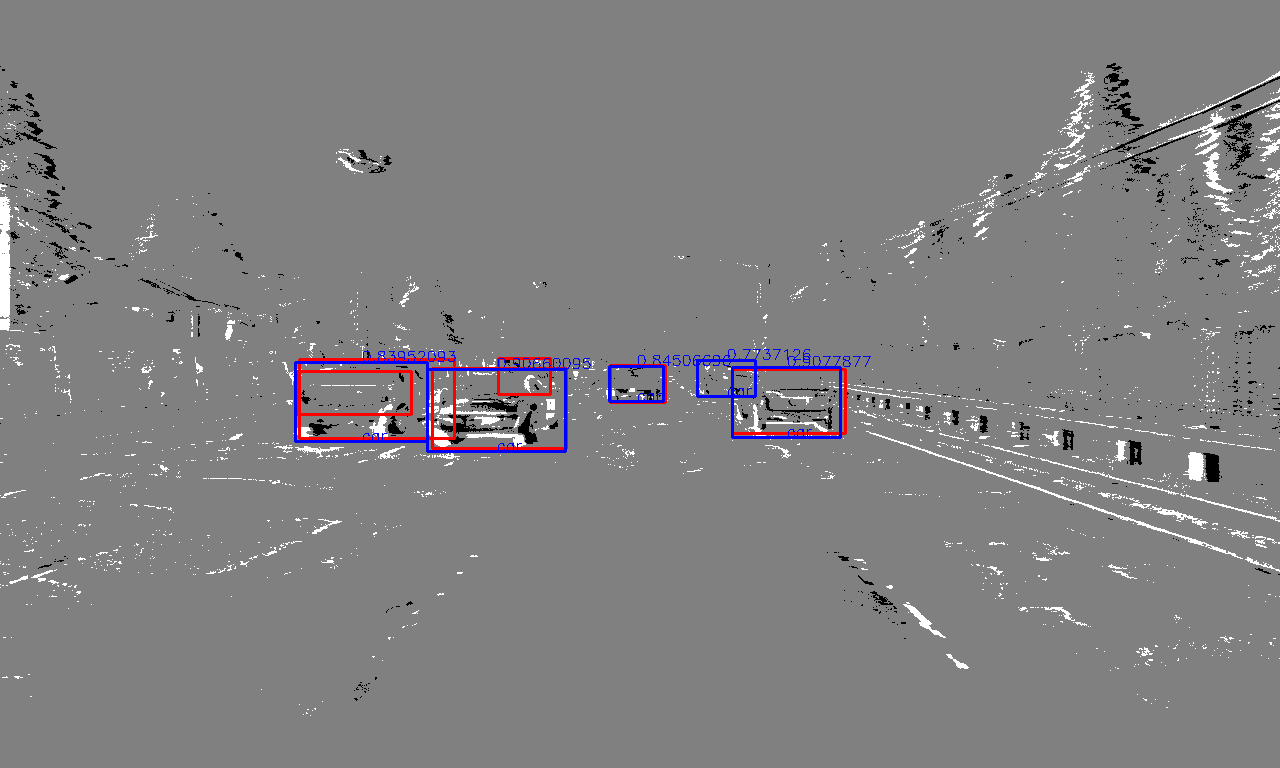}}
  \end{subfigure}
  \begin{subfigure}[c]{0.32\textwidth}
      \centering
      \fbox{\includegraphics[width=0.95\linewidth]{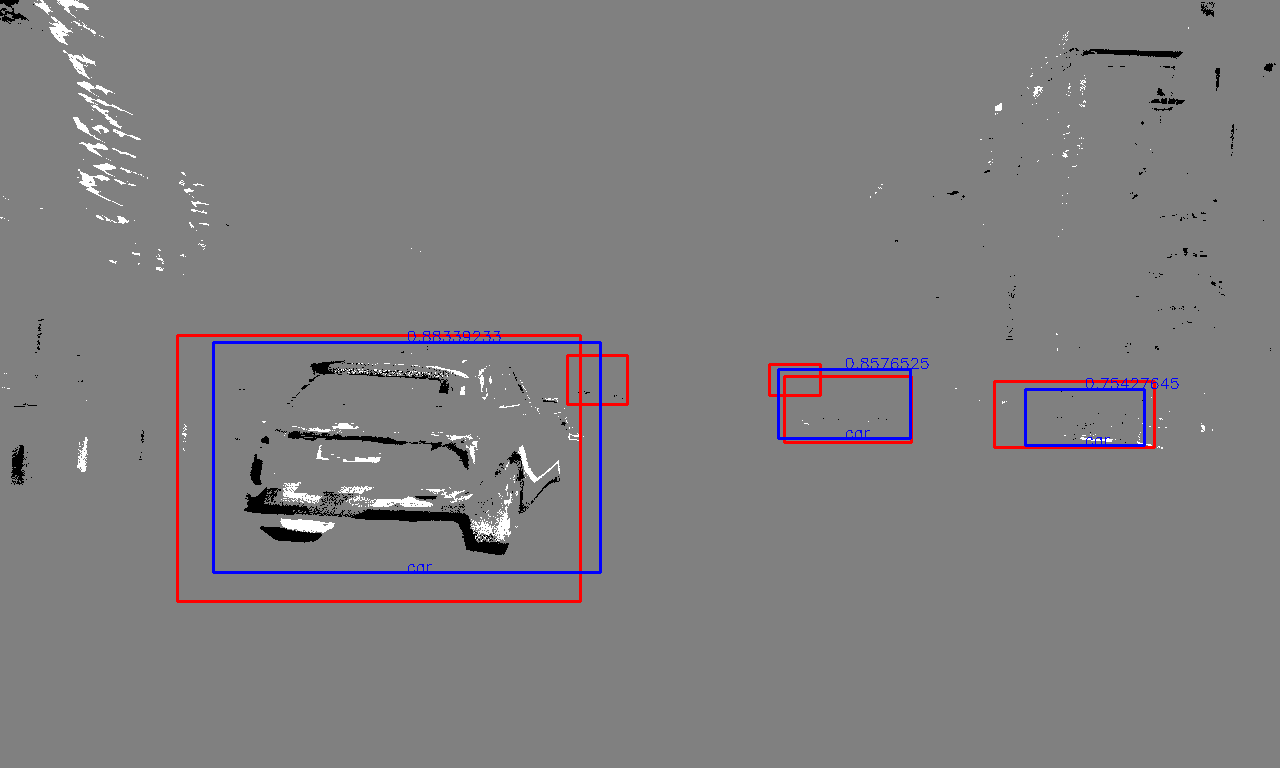}}
  \end{subfigure}
  \begin{subfigure}[c]{0.32\textwidth}
      \centering
      \fbox{\includegraphics[width=0.95\linewidth]{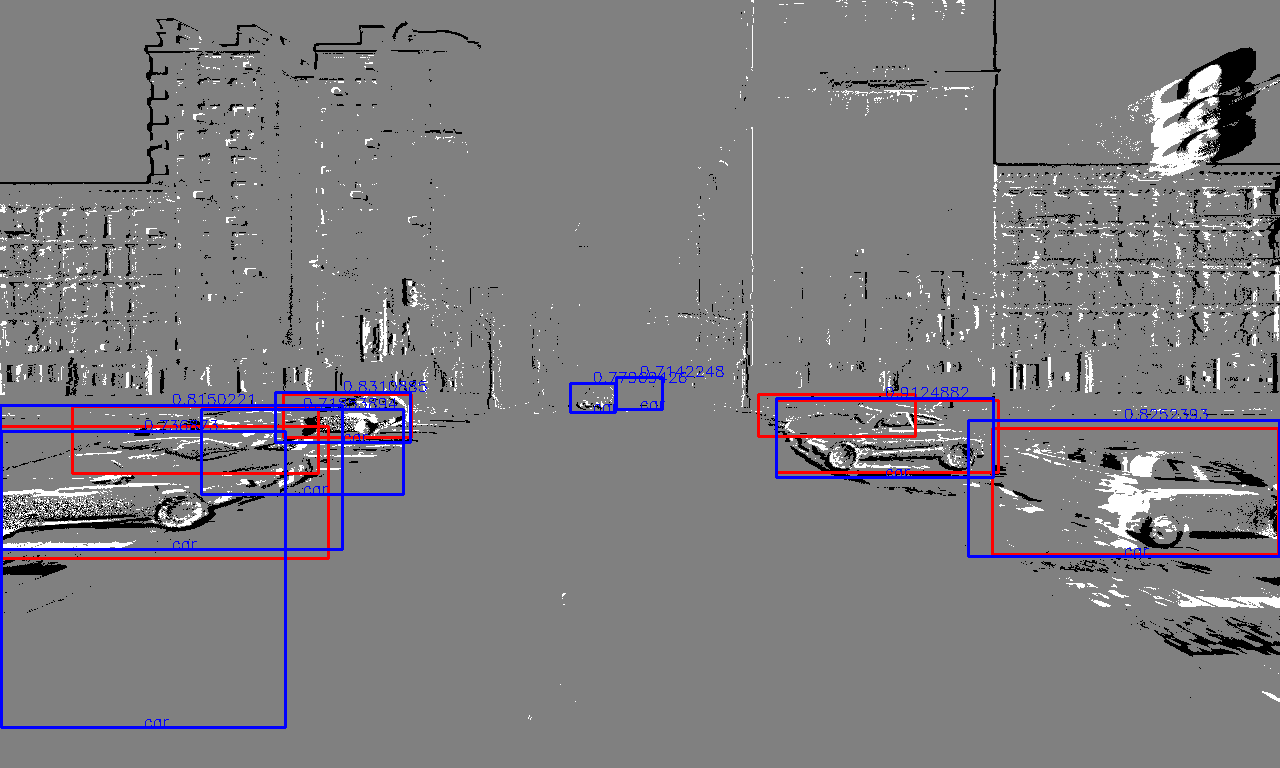}}
  \end{subfigure}
  
  \caption{Comparison of results between the different models - RVT-B and SSMS-B trained on $\mathcal{E}_{base}$ and $S_{train}$. The blue boxes indicate the predicted bounding boxes, and the red boxes represent the ground truth. Each of the three samples represents different sensor configurations to illustrate the performance across varying scenarios.}
  \label{fig:results}
\end{figure*}

We delve deeper into the outcomes in each condition, highlighting our observations, the strengths of our model, and areas that warrant further improvement. \cref{fig:results} presents a visual overview of the predictions of the models on different configurations.

\cref{table:detailed} presents the AP scores obtained by the static-trained base model and our proposed expanded model for the RVT \citep{gehrig2023recurrent} and SSMS \citep{zubic2024state} models for each of the configurations. We refer to the models trained only on $\mathcal{E}_{base}$ as the static-trained model, and the performance of all four models ($\mathcal{E}_{base}$- and $S_{train}$-trained RVT-B and SSMS-B) on configuration $\mathcal{E}_{base}$ as the baseline performance for each of them, respectively.

    \begin{table}[t]
        \centering
            \caption{Comparison of performance (AP scores in $\%$) for each configuration.}
            \label{table:detailed}
            \renewcommand{\arraystretch}{1.2}
            \begin{tabularx}{\linewidth}{|C|C|C|C|C|C|}
                \hline
                \multicolumn{2}{|c|}{\textbf{Train/Test}} & \multicolumn{2}{c|}{\textbf{RVT-B}} & \multicolumn{2}{c|}{\textbf{SSMS-B}} \\
                \cline{3-6}
                \multicolumn{2}{|c|}{\textbf{Config.}} & $\mathbfcal{E}_{\bm{base}}$ & $\bm{S}_{\bm{train}}$ & $\mathbfcal{E}_{\bm{base}}$ & $\bm{S}_{\bm{train}}$\\
                \hline
                \multirow{8}{=}{$\bm{S}_{\bm{test}}^{\bm{1}}$} 
                            & $\mathbfcal{E}_{\bm{base}}$ & \textbf{45.63} & 44.12 & 49.10 & \textbf{50.58} \\ \cline{2-6}
                            & $\mathbfcal{E}_{\bm{1}}$ & 44.49 & \textbf{45.94} & 51.99 & \textbf{53.90} \\
                            & $\mathbfcal{E}_{\bm{3}}$ & 23.74 & \textbf{30.23} & 26.72 & \textbf{33.01} \\
                            & $\mathbfcal{E}_{\bm{4}}$ & \textbf{45.13} & 44.72 & 48.97 & \textbf{50.55} \\
                            & $\mathbfcal{E}_{\bm{6}}$ & \textbf{44.61} & 44.13 & 48.73 & \textbf{50.05} \\
                            & $\mathbfcal{E}_{\bm{7}}$ & 31.49 & \textbf{35.61} & 34.68 & \textbf{41.31} \\ 
                            & $\mathbfcal{E}_{\bm{9}}$ & 7.32 & \textbf{17.10} & 7.08 & \textbf{17.58} \\
                            \cline{2-6}
                            & \textbf{Avg.} & 34.63 & \textbf{37.41} & 38.18 & \textbf{42.42} \\
                            \cline{1-6}                            
                \multirow{4}{=}{$\bm{S}_{\bm{test}}^{\bm{2}}$} 
                            & $\mathbfcal{E}_{\bm{2}}$ & 35.69 & \textbf{37.69} & 38.69 & \textbf{42.03} \\                  
                            & $\mathbfcal{E}_{\bm{5}}$ & \textbf{45.56} & 44.48 & 48.97 & \textbf{50.98} \\ 
                            & $\mathbfcal{E}_{\bm{8}}$ & 31.21 & \textbf{35.95} & 32.09 & \textbf{41.27} \\ 
                            \cline{2-6}
                            & \textbf{Avg.} & 37.49 & \textbf{39.37} & 39.92 & \textbf{44.76} \\
                            \cline{1-6}                            
                \multirow{3}{=}{$\bm{S}_{\bm{test}}^{\bm{3}}$} 
                            & $\mathbfcal{E}_{\bm{10}}$ & 33.92 & \textbf{40.00} & 40.68 & \textbf{48.14} \\
                            & $\mathbfcal{E}_{\bm{11}}$ & 34.22 & \textbf{37.70} & 40.54 & \textbf{43.13} \\ 
                            \cline{2-6}
                            & \textbf{Avg.} & 34.07 & \textbf{38.85} & 40.61 & \textbf{45.64} \\
                            \cline{1-6}                            
                \multirow{3}{=}{$\bm{S}_{\bm{test}}^{\bm{4}}$}
                            & $\mathbfcal{E}_{\bm{12}}$ & 34.08 & \textbf{37.30} & 35.79 & \textbf{40.86} \\
                            & $\mathbfcal{E}_{\bm{13}}$ & 27.09 & \textbf{31.03} & 30.88 & \textbf{37.26} \\ 
                            \cline{2-6}
                            & \textbf{Avg.} & 30.58 & \textbf{34.16} & 33.33 & \textbf{39.06} \\
                \hline
            \end{tabularx}
    \end{table}

\subsubsection{\textbf{Performance across Thresholds}} The threshold parameter, which governs the sensitivity of event generation, emerges as a critical factor influencing model robustness. When the threshold is set to a lower value, as in configuration $\mathcal{E}_1$, the system generates a higher number of events. Notably, this increase in event frequency does not adversely affect the performance of the static-trained model. In fact, for SSMs, the static-trained model even demonstrates improved results compared to its performance under the baseline threshold ($\mathcal{E}_{base}$). This counterintuitive outcome may be attributed to the model’s ability to leverage the richer temporal information provided by the denser event stream. Additionally, the event representation method employed plays a crucial role here. The Stacked Histogram representation with fixed temporal bins implements automatic saturation for high-density events through a clipping value. 

Furthermore, our model exhibits a consistent, albeit modest, improvement over the static-trained baseline under these conditions \ie $\sim1\%$ for RVT and $\sim2\%$ for SSMs. This suggests that the proposed training approach is capable of extracting additional value from the increased data density, even when the static model already performs well.

However, the scenario changes dramatically as the threshold increases, resulting in sparser event data (as seen in $\mathcal{E}_3$). Here, the static-trained model’s performance deteriorates significantly, with a reduction of $\sim23\%$ relative to the baseline. In contrast, our model, which was exposed to a broader range of event densities during training, demonstrates a more graceful degradation, \ie only $\sim15\%$ below baseline, thereby achieving an $\sim8\%$ performance gain over the static model in this challenging regime. This finding underscores the importance of training with diverse event densities to enhance model resilience under sparse data conditions.

The robustness of our model is further highlighted in out-of-distribution scenarios, such as configuration $\mathcal{E}_2$, where the threshold is set to an unseen value. Even in this setting, the model maintains its performance advantage, outperforming the static baseline by $\sim2\%$ in RVT and $\sim4\%$ in SSMs. This result attests to the generalization capabilities conferred by the employed training methodology.


\subsubsection{\textbf{Invariance to Refractory Period}}
Varying the refractory period does not have a significant effect on model performance across all tested configurations. The static-trained model maintains stable results, indicating that its learned representations are not particularly sensitive to this parameter. For the expanded model, the effect is similarly negligible, with only a slight decrease in RVT performance and a modest $\sim1-2\%$ improvement in SSMs. Importantly, this stability persists even when the values fall outside the training distribution, suggesting that both models possess a degree of invariance to this parameter that is desirable for practical deployment in variable hardware environments.


\subsubsection{\textbf{Performance across Fields of View}}
The field-of-view (FoV) parameter, which determines the angular extent of the sensor’s perceptual field, has a pronounced impact on detection performance. Altering the FoV fundamentally changes the geometry of the observed scene: increasing the FoV stretches and flattens object appearances for objects, and performs an indirect scaling operation - an increased field of view makes an object appear distant and therefore smaller. These perspective distortions can place test samples significantly out of the training distribution, challenging the model’s generalization ability.

When the FoV is narrowed to 45° (as in $\mathcal{E}_7$), the static-trained model’s performance drops precipitously by $\sim14\%$ compared to the baseline. In contrast, our expanded model exhibits a more moderate decline of $\sim9\%$, and the SSMs variant achieves a relative gain of $\sim7\%$ over the static model. This resilience is likely attributable to the model’s exposure to a wider range of FoV settings during training, which enables it to accommodate geometric distortions better.

At the other extreme, widening the FoV to 160° in configuration $\mathcal{E}_9$ leads to the most severe performance degradation across all settings. Nevertheless, when the training distribution includes such wide FoV examples, our model can outperform the static baseline by as much as $\sim10\%$. Even in intermediate, previously unseen FoV settings (e.g., 135° in $\mathcal{E}_8$), our model maintains a substantial performance margin, achieving up to $\sim9\%$ improvement in SSMs. These results collectively underscore the importance of training with a diverse range of geometric configurations to ensure robust performance.



    \begin{table*}[t!]
    \centering
            \caption{Complete Performance Comparison on the RVT-B Model and the SSMS-B Model. We observe that the SSMS-B Model trained on $S_{train}$ performs the best uniformly in each score metric, across all test sets.}
            \label{table:rvt}
            \renewcommand{\arraystretch}{1.2}
            \begin{tabularx}{\linewidth}{|C|C:C|C:C|C:C|C:C|C:C|}
                \hline
                \multirow{2}{=}{${\text{\textbf{Score}}}_{\bm{k}}$ ($\bm{f}$)}
                & 
                \multicolumn{2}{c|}{$\bm{AP (m_1)}$} & 
                \multicolumn{2}{c|}{$\bm{AP_{\bm{50}} (m_2)}$} & 
                \multicolumn{2}{c|}{$\bm{AP_{\bm{75}} (m_3)}$} & 
                \multicolumn{2}{c|}{$\bm{AP_{\bm{L}} (m_4)}$} & 
                \multicolumn{2}{c|}{$\bm{AP_{\bm{M}} (m_5)}$} \\
                \cline{2-11}
                & \textbf{RVT-B} & \textbf{SSMS-B} & \textbf{RVT-B} & \textbf{SSMS-B} & \textbf{RVT-B} & \textbf{SSMS-B} & \textbf{RVT-B} & \textbf{SSMS-B} & \textbf{RVT-B} & \textbf{SSMS-B} \\

                
                \hline
                $\text{\textbf{Score}}_{\bm{1}}$ $(\mathbfcal{E}_{\bm{base}}) $ & 34.6 $\pm$~14.7 & 38.2 $\pm$~16.6 & 59.7 $\pm$~17.9 & 60.9 $\pm$~19.8 & 37.4 $\pm$~19.7 & 42.8 $\pm$~21.6 & 44.0 $\pm$~18.0 & 49.6 $\pm$~20.4 & 30.0 $\pm$~13.4 & 32.0 $\pm$~14.9\\
                $\text{\textbf{Score}}_{\bm{1}}$ $(\bm{S}_{\bm{train}})$ & 37.4 $\pm$~10.7 & \textbf{42.4 $\pm$~13.1} & 63.4 $\pm$~11.3 & \textbf{65.3 $\pm$~14.2} & 40.7 $\pm$~15.5 & \textbf{47.7 $\pm$~17.5} & 47.4 $\pm$~14.1 & \textbf{53.8 $\pm$~17.4} & 32.3 $\pm$~9.3 & \textbf{36.4 $\pm$~11.4}\\
                \hline

                $\text{\textbf{Score}}_{\bm{2}}$ $(\mathbfcal{E}_{\bm{base}}) $ & 37.5 $\pm$~7.3 & 39.9 $\pm$~8.5 & 64.6 $\pm$~6.3 & 65.3 $\pm$~6.9 & 39.7 $\pm$~12.4 & 44.0 $\pm$~12.1 & 46.3 $\pm$~12.8 & 46.9 $\pm$~16.1 & 33.5 $\pm$~6.0 & 35.9 $\pm$~6.0\\
                $\text{\textbf{Score}}_{\bm{2}}$ $(\bm{S}_{\bm{train}})$ & 39.4 $\pm$~4.5 & \textbf{44.8 $\pm$~5.4} & 66.7 $\pm$~3.6 & \textbf{68.9 $\pm$~4.4} & 43.1 $\pm$~7.8 & \textbf{51.3 $\pm$~7.1} & 47.4 $\pm$~10.7 & \textbf{54.4 $\pm$~12.3} & 35.6 $\pm$~4.6 & \textbf{39.8 $\pm$~5.6}\\
                \hline

                $\text{\textbf{Score}}_{\bm{3}}$ $(\mathbfcal{E}_{\bm{base}}) $ & 34.1 $\pm$~0.2 & 40.6 $\pm$~0.1 & 62.3 $\pm$~4.5 & 66.7 $\pm$~2.7 & 34.4 $\pm$~3.7 & 45.2 $\pm$~0.0 & 39.3 $\pm$~3.3 & 54.9 $\pm$~4.4 & 27.6 $\pm$~1.5 &31.9 $\pm$~0.6\\
                $\text{\textbf{Score}}_{\bm{3}}$ $(\bm{S}_{\bm{train}})$ & 38.9 $\pm$~1.6 & \textbf{45.6 $\pm$~3.5} & 64.9 $\pm$~2.7 & \textbf{69.4 $\pm$~3.9} & 43.4 $\pm$~2.3 & \textbf{51.9 $\pm$~5.1} & 52.3 $\pm$~0.6 & \textbf{60.9 $\pm$~0.8} & 30.7 $\pm$~0.6 & \textbf{36.0 $\pm$~2.4}\\
                \hline

                $\text{\textbf{Score}}_{\bm{4}}$ $(\mathbfcal{E}_{\bm{base}}) $ & 30.6 $\pm$~4.9 & 33.3 $\pm$~3.5 & 57.5 $\pm$~2.6 & 59.9 $\pm$~0.4 & 29.7 $\pm$~9.2 & 34.2 $\pm$~7.1 & 40.8 $\pm$~14.2 & 43.2 $\pm$~16.3 & 24.5 $\pm$~0.6 & 28.1 $\pm$~4.2\\
                $\text{\textbf{Score}}_{\bm{4}}$ $(\bm{S}_{\bm{train}})$ & 34.2 $\pm$~4.4 & \textbf{39.1 $\pm$~2.5} & 62.5 $\pm$~0.0 & \textbf{64.9 $\pm$~0.6} & 34.5 $\pm$~9.4 & \textbf{41.9 $\pm$~4.9} & 43.7 $\pm$~15.3 & \textbf{49.0 $\pm$~14.3} & 28.9 $\pm$~1.5 & \textbf{33.5 $\pm$~4.1}\\
                \hline
            \end{tabularx}
    \end{table*}

\subsubsection{\textbf{Performance on Out-of-Distribution Configurations}}
When tested on joint parameter combinations ($\mathcal{E}_{10}$) where individual parameters resided within the training distribution but their inter-dependencies deviated from training patterns, our methodology exhibited moderate performance degradation: a $\sim4\%$ reduction for RVT and $\sim2\%$ for SSMs relative to baseline. This contrasts sharply with the static-trained model, which suffered disproportionately larger declines under equivalent conditions. Notably, the performance gap widens when both individual parameters and their joint configurations fall outside the training domain, as in $\mathcal{E}_{12}$, with our approach maintaining a consistent $\sim4-6\%$ advantage across metrics. These results suggest that while parameter interdependence introduces complexity, the proposed training strategy mitigates cascading errors through learned representations that decouple spurious parameter correlations.


\subsubsection{\textbf{Response to Asymmetric Positive and Negative Thresholds}}
A critical stress test evaluated model robustness to asymmetric event-generation thresholds, where positive and negative polarity triggers operated under distinct criteria ($\mathcal{E}{11}$, $\mathcal{E}{13}$), which was completely absent from all training data. The static-trained model exhibited pronounced sensitivity, with performance plummeting by $\sim11\%$ (RVT) and $\sim9\%$ (SSMs) when tested on threshold pairs independently overlapping its training distribution. Under fully novel asymmetries, degradation intensified to $18$–$20\%$ relative to baseline. In contrast, our expanded model demonstrated superior adaptability, sustaining only $\sim7\%$ (RVT) initial degradation, which increased to $13-14\%$ under extreme distributional mismatch. The systematic $4-5\%$ performance gap across all asymmetric configurations underscores the method’s capacity to generalize beyond symmetric threshold paradigms, a capability likely rooted in its exposure to diverse event-density variations during training, which fosters robust sensor-parameter invariance.


\subsubsection{\textbf{Differences between RVT and SSMS}}

In general, we observe that the SSMS model, paired with our training strategy, performs the best across all sensor scenarios and learns robust parameter disentanglement. It has a greater capability to generalize between the representations of the vast data distribution, compared to the RVT.

Notably, even the baseline performance (on $\mathcal{E}_{base}$) of the static-trained model improves by $\sim4\%$ when employing SSMS instead of RVT. For the same model, testing on $\mathcal{E}_{1}$, which has a significantly denser event representation due to its low threshold, this performance gap increases to $\sim7\%$, which our model further slightly widens. This shows that the frequency-generalizing adaptations adopted in SSMS also extend their profitability to the handling of varying event densities. However, when the data is significantly out-of-distribution, as in $\mathcal{E}_{9}$, any such significant performance gap between the models vanishes, both for the baseline and our expanded model. 


\subsubsection{\textbf{Other Metrics}}
We summarily present the results for all metrics, $m \in \mathcal{M}$, for each of the test sets, in \cref{table:rvt}. We present the collective scores $\text{Score}_{k}(f)$, instead of a detailed breakdown, as all individual metrics follow a similar pattern to the Average Precision (AP) scores, outlined and discussed at length.

\section{\uppercase{Conclusion}} \label{sec:conclusion}

In this work, we present a systematic approach towards sensor-invariant event-based object detection, addressing the fundamental challenges posed by varying sensor configurations. We accumulate an expansive simulated dataset to study the impact of varying sensor characteristics on the performance in downstream tasks. We highlight the limitations and weaknesses of a static-trained model in adapting to changes in signal distribution. By leveraging multi-source domain generalization, we demonstrated that training across diverse sensor parameters enhances the adaptability of event-based vision models, mitigating performance degradation due to fluctuating sensor characteristics. To the best of our knowledge, this is the first work to carry out such an expansive study into underlying sensor characteristics and the expansion of domain generalization into this scope of event data.

Through rigorous evaluation across a range of carefully designed test settings, we established the effectiveness of our employed strategy in expanding model robustness beyond standard configurations. Our findings underscore the significance of dataset diversity in training sensor-agnostic models, thereby paving the way for more reliable and scalable event-based perception systems. At the same time, we analyze failure points of our model, which warrant further research and improvement.

We believe that this work paves the way for effective future research focusing on developing dynamic adaptation mechanisms that allow for real-time tuning of sensor configurations based on environmental factors. 
Additionally, incorporating cross-modal fusion with RGB-based systems may further enhance generalization and improve event-camera integration into mainstream applications.

\section*{\uppercase{Acknowledgements}}

This work is a part of the project \textit{SNACE}, which is partially funded by the Volkswagen Foundation under the profile area \textit{Exploration} and the funding framework \textit{NEXT: Neuromorphic Computing}. 

\bibliographystyle{apalike}
{\small
\bibliography{bib}
}

\end{document}